\documentclass{mindlab}

\usepackage{amsmath,amsfonts,bm}

\def\eqref#1{equation~\ref{#1}}

\def\1{\bm{1}}

\DeclareMathAlphabet{\mathsfit}{\encodingdefault}{\sfdefault}{m}{sl}
\SetMathAlphabet{\mathsfit}{bold}{\encodingdefault}{\sfdefault}{bx}{n}

\usepackage{hyperref}
\usepackage{url}
\usepackage{amsfonts}
\usepackage{amsmath}
\usepackage{nicefrac}
\usepackage{xcolor}
\usepackage{wrapfig}
\usepackage{float}
\usepackage{needspace}
\usepackage{enumitem}
\usepackage{array}
\usepackage{longtable}

\DeclareTextFontCommand{\textbf}{\bfseries}

\hypersetup{
  pdftitle={Scaling Parameter-Efficient Fine-Tuning for Brain Encoding},
  pdfauthor={Josh Ying, Guanglong Sun, Nolan Ho, Kieran Liu, Rui Yang, Andrew Chen, Nikolaus Kriegeskorte, Pony Ma, Hossein Adeli}
}

\title{Scaling Parameter-Efficient Fine-Tuning\\for Brain Encoding}
\author[1,2]{Josh Ying}
\author[3]{Guanglong Sun}
\author[2]{Nolan Ho}
\author[2]{Kieran Liu}
\author[2,4]{Rui Yang}
\author[2]{Andrew Chen}
\author[1]{\mbox{Nikolaus Kriegeskorte}}
\author[2]{Pony Ma}
\author[1]{Hossein Adeli}
\affiliation[1]{Columbia University}
\affiliation[2]{Mind Lab}
\affiliation[3]{Tsinghua University}
\affiliation[4]{HKUST(GZ)}
\metadata[\faGithub\ Code]{\url{https://github.com/zfying/brain-lora}}
\metadata[\faEnvelope\ Correspondence]{\texttt{\textcolor{mindlabblue}{\{zy2559, ha2366\}@columbia.edu}}}
\metadata[\faCalendar\ Date]{October 2026}

\abstract{Brain encoding models aim to predict neural responses to stimuli. The most widely used approach fits readouts from frozen pretrained backbones. Using neural data to train large models from scratch or fully fine-tune them after task-pretraining has proven difficult because neural data is costly and limited. We show that parameter-efficient fine-tuning (PEFT) enables us to effectively adapt very large pretrained backbones for brain encoding with a small number of trainable parameters.
To make it computationally practical, we exploit the backbone's causal structure to predict multiple timepoints in a single pass instead of using sliding windows, significantly reducing feature-extraction cost.
This enables efficient fine-tuning of multimodal backbones with up to 104B parameters, the largest model adapted for brain encoding to our knowledge.
Trained with a low-rank adapter (LoRA) and a lightweight readout, our approach consistently improves brain encoding performance across nine diverse multimodal backbones. On the same \textsc{Qwen3-Omni} backbone as the prior state-of-the-art (SoTA), our method \textsc{BrainLoRA} achieves the best published performance on the Algonauts 2025 Challenge,
using only about $5\%$ as many trainable parameters.
We also show the adapted multimodal backbones remain interpretable: modality attribution recovers the canonical sensory topography that prior work often obtains by construction from separate unimodal encoders, and neural adaptation increases the contribution of audio inputs and cross-modal mixing within the backbone. 
To scale out the number of adapted models, we develop a multi-LoRA kernel that accelerates joint training of multiple LoRAs on a shared backbone. This enables an ensemble of 200+ adapted models that ranks first on the public Algonauts post-challenge leaderboard at the time of writing.
Together, our results highlight PEFT as an efficient and powerful paradigm for brain encoding.
}

\begin{document}

\maketitle

\section{Introduction}
Human brain encoding models predict neural responses to stimuli, providing computational insights into how the brain processes information.
State-of-the-art (SoTA) encoding models mostly follow a goal-driven paradigm \citep{yamins2016goal}: a large model pretrained on behavioral tasks is used as a \emph{frozen} feature extractor, and only a readout is learned from neural data \citep{adeli2025transformer,hwang2025silico,gokce2026mirage,dascoli2026tribe}. This approach leverages strong pretrained priors, but misses the opportunity to exploit the backbone's expressiveness to specialize for brain encoding.

Adapting large pretrained backbones with neural data promises to scale models to the complexity required to capture human brain computation, and to best combine neural information and behavioral task training. In practice, however, such adaptation is difficult, because neural data are costly and scarce relative to the size of contemporary models. Methods that do let neural data shape the representation have had mixed success. Training smaller models from scratch \citep{st2023brain,khosla2026higher} has not reached SoTA performance on whole-brain human benchmarks, and full fine-tuning of a pretrained backbone has been reported to underperform freezing it \citep{yang2023memory}.
Low-rank adaptation won Algonauts 2023 \citep{yang2023memory}, but only with a relatively small ViT, and it has not been tested on large multimodal models.

We argue that \textbf{parameter-efficient fine-tuning (PEFT) is a powerful yet underexplored regime} for building strong brain encoding models.
It retains the strong prior and expressive capacity of a large pretrained model \emph{and} allows limited neural data to adapt its representations through a small set of parameters (\S\ref{sec:formulation}). 
While the framework applies to any pretrained backbone, stimulus modality, or neural recording, we evaluate it in a particularly demanding setting: the CNeuroMod/Algonauts 2025 Challenge \citep{gifford2025algonauts}, which requires predicting whole-brain fMRI responses to long, naturalistic audiovisual movies and tests generalization to held-out out-of-distribution stimuli. We show below that PEFT is powerful at scale, setting a new SoTA on this benchmark while remaining efficient.

\begin{figure}[t]
\centering
\includegraphics[width=0.70\linewidth]{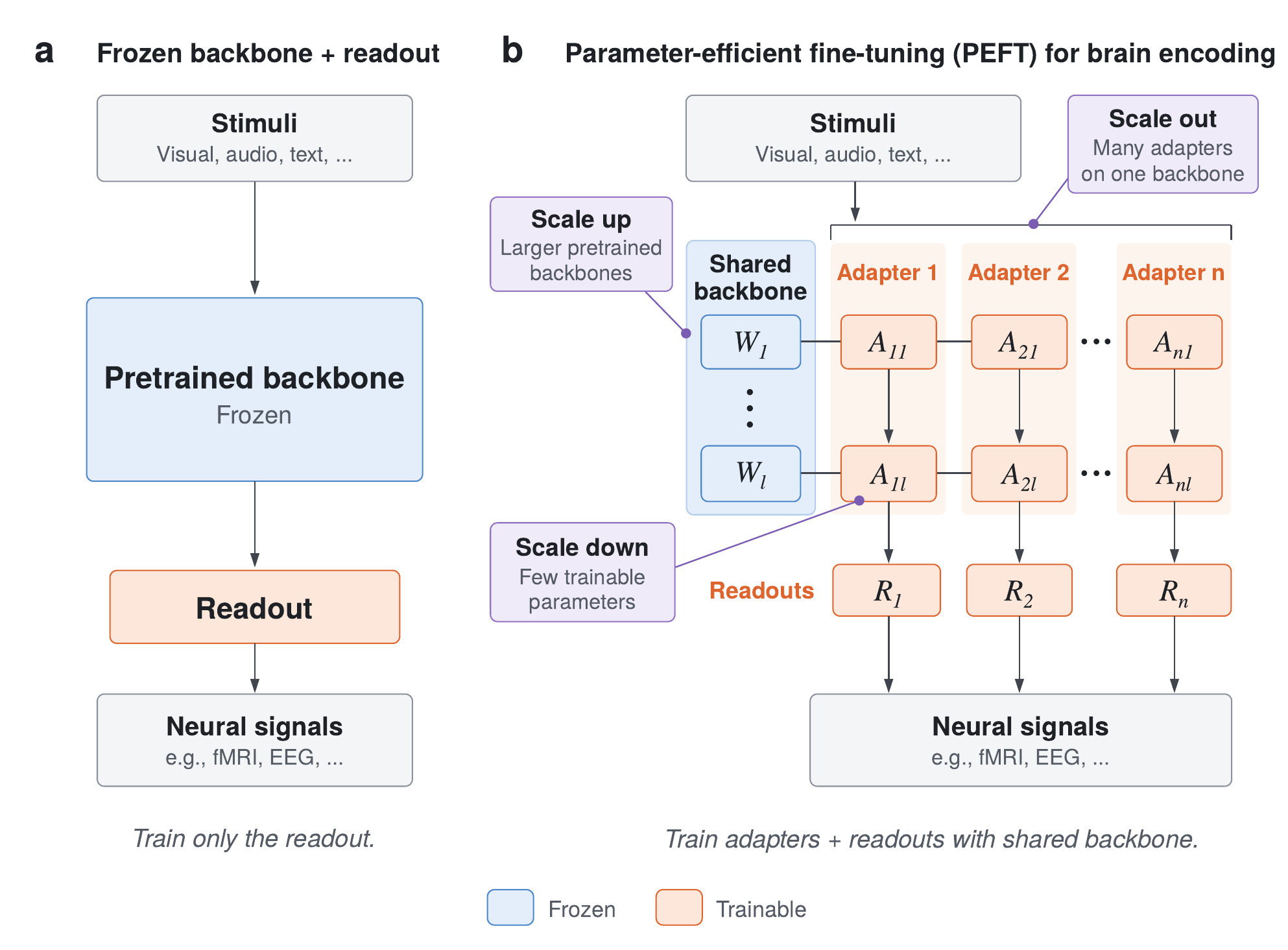}
\vspace{-15pt}
\caption{
\textbf{From frozen backbones to PEFT, and three directions to scale PEFT for brain encoding.}
\textit{(a)}~The dominant paradigm: a frozen pretrained backbone with readouts trained for brain encoding. The backbone prior is strong, but it is not specialized for brain encoding.
\textit{(b)}~We advocate using PEFT to adapt strong pretrained backbones, retaining the powerful prior while allowing flexible neural adaptation.
This suggests three complementary scaling directions: \textbf{Scale up} to stronger priors leveraging more capable backbones; \textbf{Scale down} the size of trainable adapters for brain encoding with limited data; and \textbf{Scale out} the number of adapters with different specializations.
}
\label{fig:method}
\vspace{-10pt}
\end{figure}

We first make large-scale adaptation for brain encoding computationally practical. Prior approaches typically rely on overlapping sliding windows for feature extraction, resulting in substantial redundant computation of shared stimulus context. By exploiting the backbone's innate causal masking, we encode each training window in a single forward pass and obtain representations for all timesteps simultaneously \citep{antonello2023scaling}. This reduces feature-extraction cost by approximately ${350\times}$ relative to the sliding-window protocol used by the prior SoTA at little performance cost (\citealp{gokce2026mirage}; \S\ref{sec:method}).

With adaptation made practical, we show that low-rank adapter (LoRA; \citealp{lora}) consistently improves brain encoding on \emph{all} nine multimodal backbones we test, spanning $4.7$B--$104$B parameters, the largest models ever adapted to brain encoding to our knowledge.
On the same backbone \textsc{Qwen3-Omni} as the prior SoTA \citep{gokce2026mirage}, our method \textsc{BrainLoRA} pairs LoRA with a lightweight readout. With only about $5\%$ as many trainable parameters, \textsc{BrainLoRA} sets a new SoTA among published methods (\S\ref{sec:backbones} and \S\ref{sec:main}).

Moreover, we show the adapted multimodal model remains interpretable. Prior work often identifies a region's preferred modality by determining which unimodal encoder its readout relies on. With multimodal backbones, we instead use feature attribution to identify modality contribution, which recovers the canonical topography: early visual cortex is driven by video, and auditory and lateral temporal cortex by audio. Comparison with the frozen backbone shows that adaptation reweights the contributions of the three modalities, particularly increasing the contribution of the audio inputs, and increases cross-modal mixing within the backbone (\S\ref{sec:interpret}).

Finally, to scale the number of adapted models, we develop a batched multi-LoRA kernel that trains many LoRAs jointly on one shared backbone, significantly reducing training time. This enables 200+ independently adapted models, whose ensemble ranks first on the public Algonauts post-challenge leaderboard, surpassing unpublished entries as well. We additionally show that the ensemble performance is jointly governed by ensemble size, individual model quality, and model diversity (\S\ref{sec:scaleout}).

Together, our results suggest three complementary scaling directions for future brain encoding models under the PEFT paradigm: \textbf{Scale up} the pretrained prior by leveraging increasingly capable backbones; \textbf{Scale down} the adaptation with neural data by updating only a small number of parameters; and \textbf{Scale out} by training a large number of lightweight adapters to a shared backbone with varying specializations (Fig.~\ref{fig:method}b; \S\ref{sec:formulation}). Together, PEFT turns a single powerful pretrained model into a family of specialized encoding models without paying the cost of training each model from scratch.

Our contributions are as follows:
\begin{itemize}[noitemsep, topsep=0pt, partopsep=0pt, parsep=0pt, leftmargin=20pt]
    \item \textbf{Efficiency.}
    We make PEFT on large multimodal backbones computationally practical for brain encoding (\S\ref{sec:method}), reducing feature-extraction cost by $350\times$ via single-pass extraction compared to prior SoTA \citep{gokce2026mirage} and accelerating multi-LoRA training significantly with a batched training kernel.

    \item \textbf{Effectiveness.}
    LoRA consistently improves performance over frozen backbones on all nine multimodal backbones we test (\S\ref{sec:backbones}). LoRA with a lightweight readout beats the previous published SoTA on both ID and OOD test sets using ${\approx}5\%$ as many trainable parameters
    (\S\ref{sec:main}).

    \item \textbf{Interpretability.}
    We show that large multimodal backbones for brain encoding remain interpretable: modality attribution recovers the canonical sensory topography that prior work obtains by construction from separate unimodal encoders. Comparing with the frozen backbone reveals that adaptation increases the contribution of audio inputs and the degree of cross-modal mixing in the backbone (\S\ref{sec:interpret}).

    \item \textbf{Scalability.}
    We scale the backbone up to 104B multimodal models, the largest backbones adapted for brain encoding to our knowledge, and scale out to 200+ independently adapted models (\S\ref{sec:scaleout}). The ensemble ranks first on the public Algonauts post-challenge leaderboard at the time of writing.
\end{itemize}


\section{Related work}
\label{sec:related}
We organize related works by where the neural signal enters the encoding model pipeline (see the extended related work in App.~\ref{app:related}).

\textbf{Frozen backbone, trained readout.}
Most brain encoding methods use pretrained task-optimized backbones as \emph{frozen} feature extractors and train readouts to predict neural responses \citep{naselaris2011encoding,huth2016semantic, yamins2016goal,schrimpf2021neural}. All top-ranked entries in the Algonauts 2025 challenge \citep{gifford2025algonauts} followed this paradigm \citep{gokce2026mirage,dascoli2026tribe,schad2025vibe}. Compared to more classic voxel-wise ridge regression methods \citep{naselaris2011encoding}, recent top-performing models have been using more expressive and flexible readout \citep{dascoli2026tribe,adeli2025transformer, yu2025meta}, yet the training is still confined to only the readout module.

\textbf{Backbones fine-tuned by brain data.}
Much fewer studies have instead directly adapted pretrained models using neural data. Brain-tuning of language and speech models improves not only alignment but downstream semantic tasks \citep{schwartz2019inducing,moussa2025improving,moussa2025braintuning}, and the vision analogue gains robustness and behavioral alignment \citep{li2019learning,federer2020improved}. PEFT is already used within this family: MEM uses LoRA on vision backbones and won Algonauts 2023 \citep{yang2023memory}; BrainWavLM matches a full fine-tune with LoRA adaptation \citep{vattikonda2025brainwavlm}; \citet{scholz2025stacked} uses LoRA on a 3B Llama model on Algonauts 2025 and ranked 10th. Our work follows this direction but extends it to substantially larger model scale: to our knowledge, prior models above 15B parameters have been used only as frozen feature extractors, whereas we directly adapt backbones up to 104B.

\textbf{Other ways to incorporate neural data.}
Other approaches incorporate neural signals through auxiliary alignment objectives \citep{dapello2023aligning, gokce2026multimodal}, yet shuffled-target controls sometimes reproduce much of the reported benefit \citep{guo2024limited}; through self-supervised pretraining on neural data alone \citep{caro2024brainlm,lane2026scaling}, which captures intrinsic neural dynamics rather than stimulus-referred computation \citep{zahorodnii2025neuroprobe}; through unified stimulus--brain models \citep{wu2026brainjanus}, or train the model on neural data from scratch \citep{lurz2021generalization, klindt2017neural, st2023brain, khosla2026higher}, none of which, however, are yet outperforming brain encoders built on large task-optimized feature backbones on public benchmarks \citep{schrimpf2018brain, gifford2025algonauts}.

\section{Preliminaries}
\label{sec:formulation}
Let $X$ denote a stimulus (e.g.\ video, audio, text, or a combination thereof) and $Y$ the neural response it evokes (e.g.\ fMRI, EEG/MEG, or spiking activity). An encoding model consists of a pretrained backbone $f_\theta$ followed by a readout $g_\phi$, and is trained by
\begin{equation}
\hat{Y}=g_{\phi}\big(f_\theta(X)\big),\qquad
\min_{\theta\in\Theta,\;\phi}\;\mathcal{L}\big(Y,\hat{Y}\big),
\label{eq:encoder}
\end{equation}
where $\mathcal{L}$ is a loss appropriate for the task (e.g.\ squared error or negative correlation). The readout $g_\phi$ also performs the temporal alignment between the stimulus and the neural sampling rate.

Given pretrained weights $\theta_0\in\mathbb{R}^d$, existing approaches differ in the set $\Theta$ that is optimized over:
\begin{equation}
\Theta_{\text{frozen}}=\{\theta_0\}
\;\subset\;
\Theta_{\text{PEFT}}=\{\theta_0+\Delta(\psi):\psi\in\mathbb{R}^k\}
\;\subset\;
\Theta_{\text{full}}=\mathbb{R}^{d},
\qquad k\ll d,\;\; \Delta(0)=0.
\label{eq:regimes}
\end{equation}
With a frozen backbone, only the readout is learned from neural data. Full fine-tuning and training from scratch share $\Theta_{\text{full}}$ but differ in initialization ($\theta_0$ vs.\ random), with the latter having no task-pretrained prior. PEFT for brain encoding retains $\theta_0$ and lets neural data modify the representation only through a low-dimensional parameterization $\Delta(\psi)$. This formulation is architecture-agnostic: LoRA \citep{lora}, adapter layers \citep{houlsby2019parameter}, and prompt tuning \citep{lester2021power} are all instances of $\Delta$.

The formulation suggests three complementary axes along which such models can be scaled to achieve strong brain-encoding performance in principle: \textit{Scale up} pretrained prior to large-scale, capable $\theta_0$; \textit{Scale down} the size of the adaptation, $\dim\psi$; and \textit{Scale out} to training $K$ adaptations $\{(\psi_k,\phi_k)\}_{k=1}^{K}$ on a shared $\theta_0$ and combining their predictions $\hat{Y}_k$.

\section{Method}
\label{sec:method}
\textbf{Task and data.}
We evaluate on CNeuroMod/Algonauts 2025 \citep{boyle2023courtois,gifford2025algonauts} challenge, which contains whole-brain fMRI recordings, parcellated into 1,000 regions \citep{schaefer2018local}, from subjects viewing seven seasons of \emph{Friends}, with synchronized video, audio, and transcript. We train on four subjects, validate on Season 6, and evaluate on held-out Season 7 (ID) and movies (OOD). Our main experiments use \textsc{Qwen3-Omni} \citep{qwen3omni}, the same as the best published method \citep{gokce2026mirage}, for a controlled comparison. We also conduct experiments on a diverse set of nine backbones (Tab.~\ref{tab:roster},~\ref{tab:backbones}).

\begin{wrapfigure}{r}{0.40\textwidth}
\vspace{-0pt}
\centering
\includegraphics[width=0.40\textwidth]{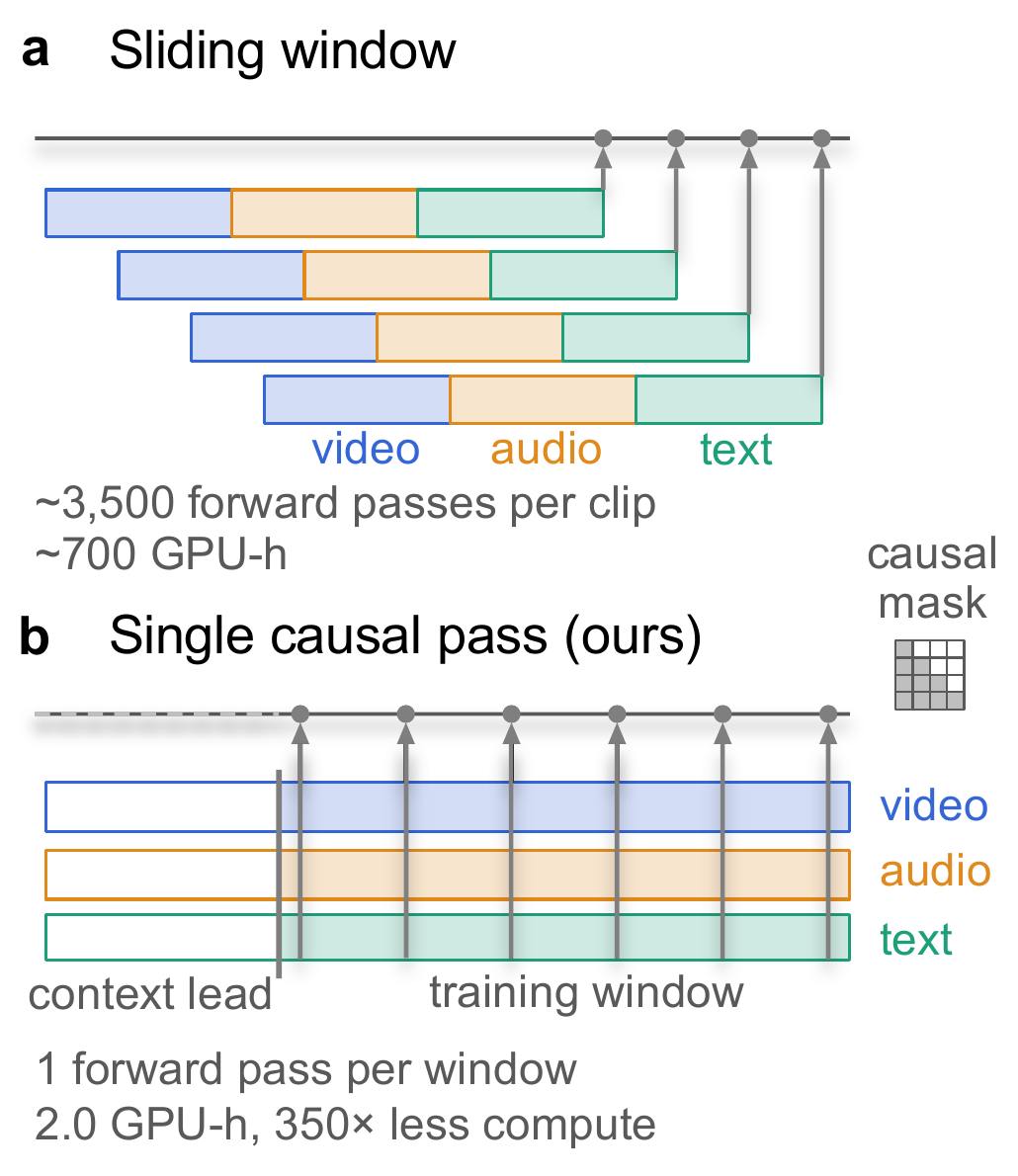}
\vspace{-20pt}
\caption{\textbf{One causally masked pass replaces a sliding window.} The baseline re-encodes an overlapping window per fMRI timestep; a single pass over the window gives features at every timestep.
}
\label{fig:extraction}
\vspace{-20pt}
\end{wrapfigure}

\textbf{Single-pass extraction.}
The prior SoTA \citep{gokce2026mirage} predicts each fMRI timestep from an overlapping sliding window, with 2{,}800 forward passes per clip (Fig.~\ref{fig:extraction}a). However, since the backbone is already causally masked, much of the repeated computation of a sliding window is unnecessary. We therefore encode each 149\,s window (100 TRs) with a single forward pass and extract representations for all timesteps simultaneously (Fig.~\ref{fig:extraction}b), an approach more common in language encoding models \citep{antonello2023scaling}.

This single-pass design reduces feature-extraction cost by ${\approx}350\times$ relative to the sliding-window protocol of \citet{gokce2026mirage} (from about 700 to 2 GPU-hours on A800), making fine-tuning through the backbone computationally affordable. Under a frozen readout, single-pass features incur only a small performance drop, which we attribute mainly to missing stimulus context at the start of each window (Tab.~\ref{tab:input-design}). We thus incorporate a 60s \emph{context lead} that largely recovers it (Fig.~\ref{fig:within-window}). With LoRA fine-tuning, our configuration surpasses the original sliding-window baseline (App.~\ref{app:input}).

The modality encoders preceding the decoder are locally bidirectional, over 8s of audio blocks and 1s of visual patches. To assess the effect of this limited non-causality, we re-encode strict prefixes of each input and compare representations at shared timesteps (App.~\ref{app:prefix}). We observe small differences, which are mainly attributable to length-dependent bf16 MoE numerics, instead of leakage of future stimuli \citep{hedefeating}.

\textbf{PEFT for Brain Encoding.} For our main method on \textsc{Qwen3-Omni}, we insert rank-16 LoRA into the attention modules through layer 32 and truncate the decoder at this layer, introducing 8.91M trainable parameters. The adapters and readout are jointly optimized for one epoch. For hyperparameter tuning, see App.~\ref{app:hparams}. We also show that other PEFT methods like $\text{IA}^3$ \citep{ia3} and adapter layers \citep{houlsby2019parameter} show similar effects (App.~\ref{app:peft}).

\textbf{Readout.}
We use linear or cross-attention \citep{adeli2025transformer} readouts. Backbone features are pooled into one vector per modality at each 2\,Hz timestep and aggregated into eight log-spaced temporal-lag blocks aligned with the fMRI sampling rate. A final linear projection maps to 1,000 brain parcels, with 52.3M trainable parameters. We pool the original ${\sim}19.3$k tokens to 894 vectors per input window. As an alternative, we replace the fixed temporal-lag aggregation with per-TR cross-attention over an expanded set of ${\sim}3.2$k tokens, adding only 0.2M parameters. Ablations in \S\ref{sec:ablation} attribute the improvement of cross-attention over linear readout primarily to the access to the richer token set.

\textbf{Closed-form re-fit.}
After joint optimization with SGD, we re-fit the linear readout $W$ in closed form. Forward passes accumulate subject-specific Gram matrices, after which $W$ is solved using parcel-wise ridge regression following \citet{nunezelizalde2019voxelwise}. As shown in \S\ref{sec:ablation}, the improvement primarily arises from the closed-form re-estimation rather than the parcel-specific regularization.

\section{Experiments}
\label{sec:experiments}
\begin{wrapfigure}{r}{0.48\textwidth}
\vspace{-25pt}
\centering
\includegraphics[width=0.46\textwidth]{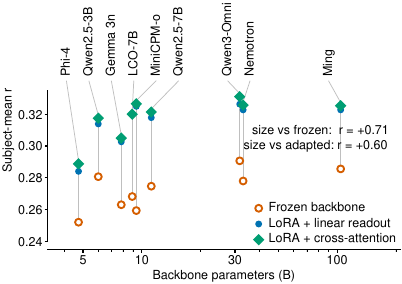}
\vspace{-5pt}
\caption{\textbf{Adaptation improves performance on nine tested backbones.} All backbones share the same training configuration except for the feature extraction layer (App.~\ref{app:backbones}). LoRA improves the performance consistently.}
\label{fig:backbone}
\vspace{-30pt}
\end{wrapfigure}

We use the mean Pearson correlation across the four subjects as the primary fMRI prediction metric. The seed-to-seed variability is small: methods whose mean correlations differ by less than 0.00077 cannot be reliably separated (LoRA+linear with 15 seeds; App.~\ref{app:protocol}). We therefore support each method comparison using seed variations whenever multiple runs are available, or for two single models, a paired bootstrap over parcels, which tests whether the performance difference holds across the cortex.

\subsection{Adapter gains across backbones}
\label{sec:backbones}

We find that PEFT significantly improves brain-encoding performance across all nine multimodal backbones tested, spanning $4.7$--$104$B parameters, compared to the frozen backbone baseline (Fig.~\ref{fig:backbone}). LoRA improves every backbone by $0.032$ to $0.066$ on linear readouts, while cross-attention yields an additional $0.003$ gain on average on all nine models.
The adaptation improves prediction performance for every parcel, with the largest gains observed in parietal and visual areas (App.~\ref{app:anatomy}). We additionally show that other PEFT methods (IA$^3$ and adapter layers) also improve over the frozen backbone significantly (App.~\ref{app:peft}).

The encoding performance and backbone size are only weakly related ($r{=}0.71$ frozen; $r{=}0.60$ adapted), as also observed in prior work on neural predictivity in large models \citep{gokce2026multimodal}.
We use \textsc{Qwen3-Omni} for the remainder of the paper, as it achieves the best performance and is also the backbone of the prior SoTA \citep{gokce2026mirage}, enabling a controlled comparison (App.~\ref{app:backbones}).

\subsection{A small adapter beats a large readout}
\label{sec:main}

As shown in Tab.~\ref{tab:main}, the previous single-model SoTA, MIRAGE \citep{gokce2026mirage}, uses a frozen backbone plus a 1.1B-parameter transformer-based readout. In contrast, our LoRA+linear model trains only 61.2M parameters (${\approx}5\%$ of MIRAGE) and already outperforms it on both ID and OOD test sets (Fig.~\ref{fig:efficiency}).
Replacing the linear readout with cross-attention, our best \emph{single model} method, \textsc{BrainLoRA}, further improves performance to 0.327 on ID and 0.228 on OOD, exceeding the published 15-model MIRAGE ensemble on ID and matching it on OOD, setting a new SoTA on the Algonauts challenge. Ensembling a population of 200+ models further reaches 0.343 on ID and 0.248 on OOD, achieving the highest score on the public leaderboard at the time of writing (\S\ref{sec:scaleout}). Selecting a subset of 15 models still maintains strong performance.

Full fine-tuning of all $20.25$B parameters below the feature extraction layer reaches $0.323$ (linear) and $0.330$ (cross-attention) on validation, on par with or slightly below the corresponding LoRA configurations, with $330\times$ more trainable parameters. The held-out test sets show the same patterns, with a larger decrease on OOD (Tab.~\ref{tab:main}). Full fine-tuning thus does not outperform LoRA at this data scale, consistent with prior works \citep{yang2023memory, zhang2024scaling, moussa2025braintuning, vattikonda2025brainwavlm}.

\begin{table}[t]
\caption{\textbf{Main results on CNeuroMod/Algonauts 2025.} Validation is on Friends season 6; ID and OOD are the challenge server's scores on held-out Friends season 7 and OOD movies.
Two rows within $0.00077$ are not separated by seed variability
(App.~\ref{app:protocol}). Rows marked $^\dagger$ are cited authors'
own numbers quoted from the papers. Our method \textsc{BrainLoRA} achieves a new state of the art.}
\label{tab:main}
\centering
\footnotesize
\setlength{\tabcolsep}{4pt}
\begin{tabular}{lrrrr}
\toprule
& Trainable & \multicolumn{3}{c}{Pearson $r$} \\
\cmidrule(lr){3-5}
Model & params & val & ID & OOD \\
\midrule
\multicolumn{5}{l}{\emph{Single model}} \\
TRIBE$^\dagger$\,\citep{dascoli2026tribe} & -- & -- & 0.303 & 0.196 \\
MIRAGE$^\dagger$\,\citep{gokce2026mirage} & 1.1B & 0.319 & 0.310 & 0.217 \\
Linear + frozen backbone & 52.3M & 0.291 & 0.287 & 0.187 \\
Linear + full finetuning & 20.3B & 0.323 & 0.320 & 0.217 \\
\textsc{BrainLoRA} (Linear) & 61.2M & 0.326 & 0.322 & 0.222 \\
Cross-attn + frozen backbone & 52.5M & 0.312 & 0.307 & 0.212 \\
Cross-attn + full finetuning & 20.3B & 0.330 & 0.326 & 0.225 \\
\textsc{BrainLoRA} (Cross-attn) & 61.4M & \textbf{0.331} & \textbf{0.327} & \textbf{0.228} \\
\midrule
\multicolumn{5}{l}{\emph{Ensemble}} \\
TRIBE$^\dagger$, 1000 models \citep{dascoli2026tribe} & -- & -- & 0.320 & 0.215 \\
Vibe$^\dagger$\,\citep{schad2025vibe} & $20{\times}$329M & -- & 0.323 & 0.212 \\
MIRAGE$^\dagger$, 15 models \citep{gokce2026mirage} & $15{\times}$1.1B & 0.335 & 0.323 & 0.227 \\
\textsc{BrainLoRA}, best 15 models & $15{\times}$61M & 0.347 & 0.341 & 0.247 \\
\textsc{BrainLoRA}, 261 models & $261{\times}$61M & \textbf{0.348} & \textbf{0.343} & \textbf{0.248} \\
\bottomrule
\vspace{-12pt}
\end{tabular}
\end{table}

\textbf{The adaptation transfers to unseen subjects and a new dataset.}
The learned adaptation generalizes beyond both the subjects and the dataset used for training, with gains essentially matching those observed in-distribution.
When trained on three subjects and evaluated on a held-out fourth subject with the LoRAs frozen and only the readout re-fitted, it retains $97\%$ of the improvement observed for subjects included in LoRA training (App.~\ref{app:loso}).
The same adapter and readout also transfer \emph{zero-shot} to the Naturalistic Neuroimaging Database (NNDb; \citealp{aliko2020nndb}), collected at a different site with a different scanner and repetition time, and with subjects viewing films. The adapter is applied without any further training on NNDb, including the readout. Even so, it yields a gain of $+11.1$ percentage points of the noise ceiling, on par with the $+10.7$-point fine-tuning gain on our validation set (App.~\ref{app:nndb}). With only minutes of a new subject's data, solving its readout as a correction to the population readout improves the transfer performance (App.~\ref{app:fewshot}). The adapter thus captures stimulus representations that generalize to other subjects or datasets.

\subsection{Ablating the model components and the input design}
\label{sec:ablation}

\textbf{Model components.}
Tab.~\ref{tab:ablation} quantifies the contribution of each component in our main configuration. The LoRA adapter provides the largest improvement over the frozen-backbone baseline, increasing performance by $+0.0356$. This improvement disappears when the adapter is trained on fMRI shifted out of alignment with the stimulus (App.~\ref{app:shift}). The closed-form re-fit contributes $+0.0061$ of the adapted model's score, an effect larger than those obtained from any of the other changes to the training procedure we evaluated. This improvement pattern is consistently observed across all tested configurations.

\begin{wraptable}{r}{0.42\textwidth}
\vspace{-12pt}
\caption{\textbf{Ablation.} LoRA r16+linear readout, 60s context lead, closed-form re-fit. The first block builds our main configuration; the rest vary one choice from the main configuration. Rows within the 0.00077 seed spread are not separable.}
\label{tab:ablation}
\centering
\footnotesize
\setlength{\tabcolsep}{6pt}
\begin{tabular}{lr}
\toprule
Configuration & s06 \\
\midrule
Frozen backbone & 0.2907 \\
+ LoRA r16 \textbf{(ours)} & 0.3263 \\
\quad without closed-form re-fit & 0.3202 \\
\midrule
Rank 32 & 0.3268 \\
Gated-MLP readout & 0.3269 \\
Cross-attention readout & 0.3310 \\
\midrule
No audio & 0.3026 \\
\textsc{Qwen3.8-27B} backbone & 0.3191 \\
\textsc{Gemma-3-27B} backbone & 0.3029 \\
\bottomrule
\end{tabular}
\vspace{-12pt}
\end{wraptable}

\textbf{Input design.}
Tab.~\ref{tab:input-design} in App.~\ref{app:input} evaluates our input-design choices incrementally, under the MIRAGE readout heads or linear heads, on frozen or adapted backbones. Replacing the sliding windows with a single causally masked forward pass over the full clip reduces performance by only $0.0038$ under the \emph{frozen head}, at two orders of magnitude less compute. Restricting the causal pass to the $149$\,s (100 TR) training window causes a further $0.0046$ drop, concentrated at the start of each window (Fig.~\ref{fig:within-window}), which a context lead largely recovers, with the benefit saturating at $60$\,s. We thus adopt $60$\,s for the final configuration (App.~\ref{app:input}).

The input design choices affect the frozen and adapted cases differently. With frozen readout heads, performance decreases as input context is removed, since the readout is trained for a different representation. However, once the backbone and readout are adapted to the new input configuration, we can recover the lost performance.
Thus, the computational efficiency of single-pass encoding does not come at a cost to performance once the backbone is allowed to adapt.

App.~\ref{app:negatives} summarizes null results observed during exploration. Applying cross-attention to the same pooled feature set yields no measurable improvement, indicating that the gains from the cross-attention readout arise primarily from access to a larger token set with the same number of trainable parameters. Varying the training subsets likewise does not produce useful ensemble diversity. Moreover, several apparent improvements, including increasing the LoRA rank and replacing the linear readout with a gated MLP, disappear after the closed-form re-fit (App.~\ref{app:hparams}).

\subsection{Interpreting the multimodal backbone}
\label{sec:interpret}

Encoding models can be used to ask \emph{which} stimulus information is processed, and therefore drives the prediction of neural activity, in a given brain region. Prior work often answers this by construction, using separate unimodal backbones so that a region's preferred modality or semantic selectivity can be read off directly from the readout by the weights or the attention patterns \citep{adeli2025transformer, dascoli2026tribe}. A single multimodal backbone, however, lacks this explicit separation, with information from different modalities merging early in the backbone. We show that feature attribution methods can estimate modality contribution and examine whether the adapted model recovers the expected cortical topography.

\textbf{Design.} For each window, we compute exact Shapley values over the three input modalities, $\{\text{video},\allowbreak\ \text{audio},\allowbreak\ \text{text}\}$. A modality is ablated by replacing its input embeddings with a content-free mean embedding while preserving token count, position, and temporal alignment \citep{lundberg2017unified}. Therefore, only stimulus \textit{information} is removed. Evaluating all eight ($2^3$) modality subsets yields an exact additive decomposition of each modality's contribution to encoding performance.
The same decomposition can also tell us through which token positions each modality's information reaches the readout. At every timestep, the readout takes three hidden states, one for each modality, at the truncated layer. After many layers of attention, however, each of these states can carry information from all three input modalities, so audio information, for example, may reach the readout through a video token. To separate these routes, we apply the Shapley decomposition to each of the three hidden states separately. This yields a $3 \times 3$ attribution matrix whose entry $(i, j)$ is the contribution of input modality $j$ that is read out at modality $i$'s token position by the readout.

\begin{figure}[t]
\centering
\includegraphics[width=\linewidth]{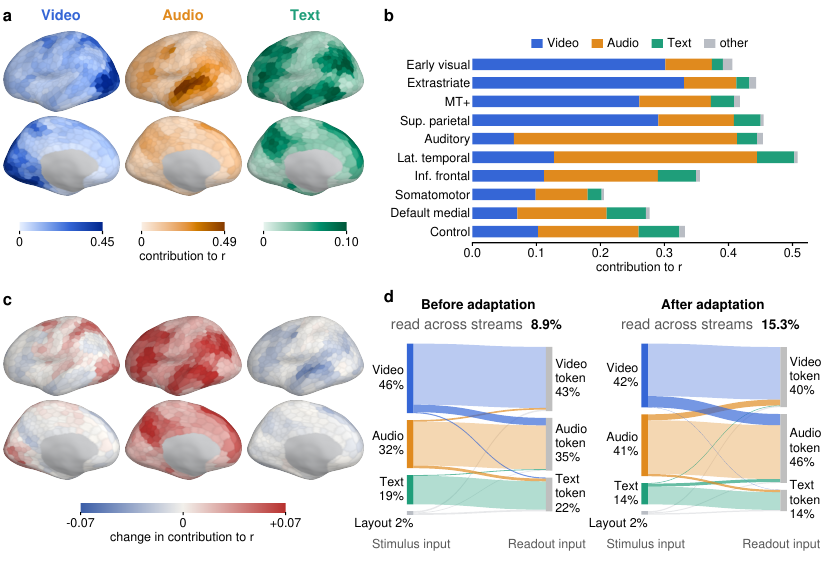}
\vspace{-30pt}
\caption{\textbf{Modality attribution recovers the canonical topography, and adaptation reweights and mixes the modalities.}
\emph{(a)} Contribution to encoding accuracy of the video, audio, and text content, on our LoRA+linear model, left hemisphere (lateral above, medial below).
\emph{(b)} Modality contribution by region. The Shapley terms from the three modalities and a small layout residual sum to the region's $r$.
\emph{(c)} Change in modality contribution from the frozen to the adapted model. Audio contribution increases the most, while text falls.
\emph{(d)} Share of the prediction carried from each stimulus modality (left) to the backbone's output at each modality's tokens (right), which is the readout's input. Crossing ribbons are cross-modal mixing within the backbone, which nearly doubles after adaptation.}
\vspace{-10pt}
\label{fig:modality}
\end{figure}

\textbf{The attribution recovers the canonical topography.}
The resulting attribution maps recover the canonical cortical organization (Fig.~\ref{fig:modality}a,b, App.~\ref{app:modality}): Early visual cortex is predominantly driven by video ($72.5\%$), auditory cortex by audio ($72.9\%$), and lateral temporal cortex by audio ($60.3\%$) rather than text ($12.5\%$). Text contributes most strongly to predicting the activity in the default-mode ($21.6\%$) and control regions. These patterns are highly robust: 1) all regional contrasts have bootstrap intervals excluding zero, 2) the attribution maps are consistent under a donor-window baseline ($r{=}0.994$)
, and 3) two independently trained seeds yield highly consistent contribution maps ($r{=}0.996$--$1.000$).

\textbf{Adaptation shifts modality attribution toward audio.}
To see how adaptation changes modality attribution, we compare the frozen baseline with the adapted one. The spatial maps are largely unchanged: each modality's contribution map correlates at $0.963$--$0.990$ between the frozen and adapted models. What changes is the relative weighting of the three modalities (Fig.~\ref{fig:modality}c). Averaged over parcels, the audio contribution rises by $39\%$ and video by $5\%$, while text falls by $17\%$. The dominant modality changes in $249$ parcels, all toward audio, with $246$ of them switching from video- to audio-dominated and none the other way. Consistent with this, removing audio hurts the performance more after adaptation (App.~\ref{app:modality}). This shift is robust across seeds, which agree on the dominant modality for $97.2\%$ of parcels.

\textbf{Adaptation increases cross-modal mixing.}
Feature attribution also reveals the cross-modal contributions within the multimodal backbone: $15.3\%$ of the prediction is attributed to information read out at another modality's token positions, with the largest cross-modal contribution coming from video information at audio-token positions ($7.5\%$). Yet the frozen backbone reads only $8.9\%$ across modalities. Adaptation thus greatly increases the cross-modality mixing in the backbone (App.~\ref{app:modality}).

\subsection{Training a population of independently adapted models}
\label{sec:scaleout}

\begin{figure}[t]
\centering
\includegraphics[width=\linewidth]{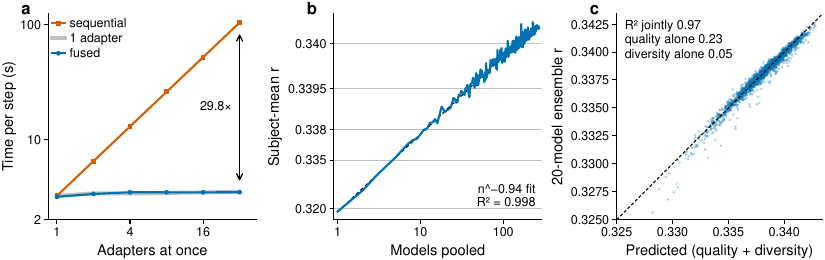}
\vspace{-20pt}
\caption{\textbf{Scaling out: training a population of adapted models.}
\emph{(a)} Training-step time for $N$ rank-16 adapters on a shared backbone: one fused batched gather-matmul vs.\ $N$ sequential runs. The fused step matches the cost of a single adapter on the same batch (App.~\ref{app:efficiency}).
\emph{(b)} Ensemble performance vs.\ ensemble size, averaged over random subsets of the 261 adapted models, which is nearly a straight line in the log-log space (App.~\ref{app:roster}).
\emph{(c)} Performance of 2{,}000 random 20-model ensembles vs.\ the performance predicted linearly on two properties of their members: mean single-model accuracy (quality) and mean pairwise residual correlation (diversity). Neither factor predicts ensemble accuracy alone ($R^2$ 0.23 and 0.05), but together they predict the performance well ($R^2{=}0.97$).
}
\vspace{-10pt}
\label{fig:scaleout}
\end{figure}

\textbf{Efficient multi-adapter training.}
To efficiently train many adapters, we develop a specialized multi-LoRA training kernel. Since LoRA modules are small and structurally identical, the routed updates for $N$ adapters can be fused into a single batched gather--matrix multiplication, enabling parallel training of $N$ adapters together (Fig.~\ref{fig:scaleout}a). The speedup arises from improved hardware utilization rather than reduced FLOPs: at $N{=}32$, the fused step costs only $1.08$x a single-adapter step, yielding a $29.8$x speedup over sequential training. End-to-end speedups are smaller because of longer sequences and larger batches.
See App.~\ref{app:efficiency} for details.

\textbf{Large-scale ensemble performance.}
We train 261 independently adapted models spanning different hyperparameters, readout architectures, and backbones, and aggregate with a parcel-wise minimum-variance ensemble combiner fitted on out-of-fold predictions (App.~\ref{app:roster}). Ensemble accuracy reaches 0.3483 on validation, 0.3428 on ID, and 0.2483 on OOD, achieving the top score on the public post-challenge leaderboard at the time of writing, outperforming unpublished submissions as well (Tab.~\ref{tab:main}). All reported models were selected using validation performance only. A carefully selected 15-model subset reaches 0.3465 on validation, within 0.0018 of the full ensemble, at one seventeenth of the inference cost. Both the subset and its weights are chosen on the validation set (App.~\ref{app:protocol}).

To characterize the effect of ensemble size, we score random subsets of the 261-model pool at every size $n$ and fit their mean accuracy with  $m(n)=m_\infty-An^{-b}$, which results in $b\approx0.94$ ($R^2{=}0.998$; Fig.~\ref{fig:scaleout}b).
An exponent near 1 is consistent with ensemble theory: if members' residuals have mean pairwise correlation $\rho$, the ensemble's residual variance is $\sigma^2[\rho+(1-\rho)/n]$ \citep{ueda1996generalization, wood2023unified}, whose second term shrinks as $1/n$. The first term does not shrink with size, which sets the ceiling performance $m_\infty$. The ensemble ceiling thus depends on how correlated the members' errors are ($\sigma^2\rho$), and hence on the ensemble's composition rather than its size alone.

We next examine, at a fixed pool size, what ensemble properties influence the performance. For random 20-model subsets, mean residual correlation and mean single-model performance explain 5\% and 23\% of the variance in ensemble score individually, but explain 97\% of the variance jointly (Fig.~\ref{fig:scaleout}c; \citealp{krogh1995ensembles, kuncheva2003measures, douven2026condorcet}). These factors are highly correlated in our pool ($r{=}0.74$), as models with more diversity tend to be weaker individually. Therefore, separate coefficients of the fit should not be read as the effect of changing one property while holding the other fixed.

Useful diversity arises primarily from changes in model architecture rather than minor training variations. Models with different readout architectures or backbones show substantially lower residual correlation with the rest of the pool. In contrast, changes in seed, LoRA rank, layer choice, or training data subset provide little diversity (Fig.~\ref{fig:seedgeom}, App.~\ref{app:seedgeom}).

\section{Discussion}
\label{sec:discussion}
\textbf{Three axes for scaling PEFT for brain encoding.}
Together, our results suggest three complementary scaling directions for future brain encoding models: \textbf{1) Scale up} the pretrained prior by leveraging increasingly capable backbones; \textbf{2) Scale down} the adaptation by updating only a small number of parameters; and \textbf{3) Scale out} by training a large number of lightweight adapters to shared backbones with various inductive biases \citep{lab2026scaling}. Together, PEFT turns a single powerful pretrained model into a family of specialized encoding models without training each backbone from scratch.

\paragraph{Toward a brain foundation model.}
Our experiments do not constitute a brain foundation model, but scaling along the three directions discussed above could lead to one. A large pretrained backbone could use a single shared adapter trained on neural data spanning tasks, datasets, subjects, and even species. In this work, we only scale out the number of adapters for ensemble. However, future work could train a separate adapter for each task, dataset, subject, or even ROI with a hierarchical structure or with meta-learning.
Together, this hierarchical approach and the three scaling directions we discuss above can create a brain foundation model with efficient specializations for different downstream tasks.

\textbf{Limitations.}
Our models are trained on one benchmark with four subjects, all watching audiovisual narrative. The adapters transfer to a held-out subject (App.~\ref{app:loso}) and to a second dataset (App.~\ref{app:nndb}), but neither test leaves the domain of narrative film. We leave the joint training on multiple datasets and modalities to future work.

\bibliographystyle{plainnat}
\bibliography{paper}

\clearpage
\beginappendix
\section{Extended related work}
\label{app:related}

The full version of \S\ref{sec:related}, by family: \emph{where does the
neural signal enter the encoding model pipeline?}

\paragraph{Frozen backbone, trained readout.}
The dominant paradigm freezes a model pretrained on a behavioral or self-supervised task and fits a readout from its hidden states to neural responses: the lineage running from voxelwise ridge regression \citep{naselaris2011encoding,huth2016semantic,nunezelizalde2019voxelwise} through task-optimized network comparisons \citep{yamins2016goal,schrimpf2021neural,caucheteux2022brains} to today's multimodal encoders. Every leading entry in the Algonauts 2025 challenge leaderboard \citep{gifford2025algonauts} followed this paradigm, including MIRAGE \citep{gokce2026mirage}, which we build on directly, TRIBE \citep{dascoli2026tribe} and VIBE \citep{schad2025vibe}. A retrospective report argued that architectural choices mattered less than ensembling \citep{scotti2025insights}. Readout design has been studied on its own terms \citep{adeli2025transformer,lurz2021generalization}, and a \emph{larger} readout does not necessarily lead to higher accuracy: an ROI-query transformer readout is ${\sim}$37M against classical ridge's ${\sim}$1{,}200M, yet it performs better \citep{adeli2025transformer}. Concurrent work \citep{zhang2026platonic} reports a strong OOD score on the Algonauts 2025 leaderboard using a pool of frozen-backbone models with linear readouts, whose per-participant, per-film parameters were selected using public leaderboard scores. By contrast, we selected all models and ensemble weights on validation data alone (App.~\ref{app:protocol}).

\paragraph{Backbones fine-tuned by brain data.}
A second family updates the stimulus model's own weights from neural recordings. Inducing brain-relevant bias in language models \citep{schwartz2019inducing,toneva2019interpreting} leads to brain-tuning of speech models, which improves not only alignment but also downstream semantic tasks \citep{moussa2025improving,moussa2025braintuning}. The vision analogue trains with neural regularization and gains robustness and behavioral alignment \citep{li2019learning,federer2020improved,safarani2021towards,lu2024achieving}. Our methods fall under this family, and parameter-efficient adaptation is already used within it: BrainWavLM brain-tunes with LoRA and matches a full fine-tune \citep{vattikonda2025brainwavlm}, and rank-8 LoRA at $0.625\%$ of parameters suffices for cross-participant brain-tuning \citep{moussa2025braintuning}; MEM won the Algonauts 2023 challenge by adapting a vision backbone with LoRA \citep{yang2023memory}; recent work RABBiT achieves SoTA on audio cortex prediction by adapting an audio backbone \citep{moussa2026rabbit}. Although several works have explored PEFT for brain encoding, this remains a relatively underexplored paradigm in the literature. What is new about our work is mainly the \textbf{scale}: the largest backbones whose weights have been updated by brain data are around 10B \citep{scholz2025stacked,wu2026brainjanus}, and models above 15B appear in this literature only as \emph{frozen} feature extractors.
Concurrent work \citep{zhang2026platonic} also trains LoRA experts inside omni-modal models of up to 104B total parameters, but supervises them with labels derived from fMRI networks rather than with the brain-encoding objective of predicting neural responses.
We adapt backbones up to 104B for brain encoding, and we test them on a whole-brain encoding benchmark rather than on a specific brain area or for downstream task transfer.

\paragraph{Joint task and brain-alignment objectives.}
A third family trains a primary objective together with an auxiliary alignment term against neural data \citep{dapello2023aligning,pirlot2022improving}. The binding constraint is that the stimuli with recordings are a tiny fraction of the task corpus, which is precisely the asymmetry parameter-efficient adaptation is built for. This family also involves some methodological warning signs: shuffled- or permuted-target controls sometimes reproduce much of the reported benefit \citep{federer2020improved,guo2024limited}, so a gain from neural supervision has to be shown against matched compute. We hold ourselves to the same standard throughout (\S\ref{sec:ablation}), and run the shifted-target control (App.~\ref{app:shift}).

\paragraph{Brain-only pretraining, aligned afterwards.}
A fourth family pretrains self-supervised on neural data alone across subjects, sessions and datasets: masked autoencoding for fMRI \citep{caro2024brainlm,dong2024brainjepa,lane2026scaling}, EEG \citep{jiang2024labram,jiang2025neurolm}, intracranial recordings \citep{wang2023brainbert,zheng2024duin} and other modalities, and attaches the stimulus afterwards. It is very hard to scale along this axis, and the recent evidence suggests that these methods capture intrinsic dynamics and connectivity rather than stimulus-referred computation: the seed work for this family reports a null result for brain-only pretraining on trait prediction \citep{lane2026scaling}, an intracranial benchmark finds such models losing to linear baselines on stimulus decoding \citep{zahorodnii2025neuroprobe}, and pretraining gains shrink under data-matched controls \citep{ye2025ndt3}.

\paragraph{Unified models over mixed tokens.}
A fifth family puts stimulus and brain tokens into one model so that encoding and decoding share parameters \citep{wu2026brainjanus,mai2026neuroflow, qian2024lea,antoniades2024neuroformer}. It is worth noting that the most recent of these methods adapts a large pretrained backbone with LoRA rather than training from scratch \citep{wu2026brainjanus}, the same direction we are going. No model in this family has been shown to beat a specialized encoder on voxelwise correlation; they aim for decoding or semantic-level proxies instead.

\paragraph{Batched multi-adapter kernels.}
Routing one batch of tokens to many adapters is a solved problem for inference: Punica's segmented gather matmul \citep{punica} and S-LoRA's paged adapter pool \citep{slora} both handle it. Because serving needs no gradients, neither system provides a backward pass. Training systems are more recent and batch per adapter, with each adapter receiving its own inputs, so every group is a full sequence \citep{plora,alto}. The per-token form of the primitive instead comes from mixture-of-experts training, where MegaBlocks pads groups to block multiples \citep{megablocks} and ScatterMoE does not \citep{scattermoe}; stock PyTorch now also ships a differentiable grouped matmul that takes a distinct matrix per group. Neither the operator nor our two design choices (gathering rather than padding, and carrying the adapter index on a grid dimension) is new.
What is specific to our method is the regime. Grouped matmuls are designed for many tokens per group, whereas a pool with one adapter per model, subject or parcel sits at the opposite extreme, where each group holds only a handful of rows. In that regime, the fused kernel's cost is nearly independent of pool size, while the alternatives' costs grow with it. At the pool sizes used in this paper, this is merely a speedup; if we were to train one adapter per parcel, our kernel would make training feasible at all (\S\ref{sec:discussion}).

\section{The full input design}
\label{app:input}

\begin{table}[t]
\caption{\textbf{From MIRAGE's sliding window to a single causal pass.}
Friends s06 subject-mean Pearson over four subjects. The sliding-window
protocol re-encodes the preceding stimulus at every timestep ($\sim$2{,}800
forward passes per clip); a single causal pass encodes each clip or window
once, and a context lead supplies the preceding stimulus that the first
timesteps of a window would otherwise lack. Step time and training cost are
for the linear readout. The sliding-window cost is the $\sim$700 GPU-hours
reported by the baseline's authors, which we reproduced on our hardware
(App.~\ref{app:efficiency}). The frozen MIRAGE head was trained at a 30\,s
lead and has no longer-lead entries; with LoRA the head was trained at 30 and
60\,s, and its re-fit re-solves the per-subject output layer in closed form.
Both LoRA column pairs show seed 0. The linear readout has no whole-clip
entries because whole-clip encoding is the limit of the context length sweep.
}
\label{tab:input-design}
\centering
\footnotesize
\setlength{\tabcolsep}{3pt}
\begin{tabular}{lrrrrrrrr}
\toprule
& Extract & \multicolumn{3}{c}{MIRAGE head (1{,}123M)} & \multicolumn{2}{c}{Linear (52.3M)} & & Train \\
\cmidrule(lr){3-5}\cmidrule(lr){6-7}
Stimulus design & (GPU-h) & frozen & LoRA & + re-fit & LoRA & + re-fit & s/step & (GPU-h) \\
\midrule
Sliding window, full-res.\ media & $\sim$700 & 0.3187 & -- & -- & -- & -- & -- & -- \\
Sliding window, $336{\times}224$ media & $\sim$700 & 0.3178 & -- & -- & -- & -- & -- & -- \\
Single pass, whole clip & 1.99 & 0.3149 & -- & -- & -- & -- & -- & -- \\
Single pass, whole clip, $336{\times}224$ & 1.99 & 0.3137 & -- & -- & -- & -- & -- & -- \\
\quad + window-local (149\,s), no lead & 1.99 & 0.3091 & -- & -- & 0.3148 & 0.3206 & 9.7 & 4.0 \\
\quad + 30\,s context lead & 1.99 & 0.3124 & 0.3214 & 0.3236 & 0.3187 & 0.3249 & 10.4 & 4.3 \\
\quad + 60\,s context lead & 1.99 & -- & 0.3220 & 0.3247 & 0.3202 & \textbf{0.3263} & 12.0 & 4.8 \\
\quad + 90\,s context lead & 1.99 & -- & -- & -- & 0.3196 & 0.3260 & 12.9 & 5.2 \\
\quad + 120\,s context lead & 1.99 & -- & -- & -- & 0.3203 & \textbf{0.3263} & 14.4 & 5.6 \\
\bottomrule
\end{tabular}

\end{table}

Tab.~\ref{tab:input-design} gives the full results behind \S\ref{sec:ablation}. Lowering the stimulus resolution to $336{\times}224$ costs only 0.0009, so all later rows use the low-resolution media. Replacing the sliding window with a single causal pass over the whole clip costs 0.0038 under the frozen MIRAGE head at original resolution (0.3187 against 0.3149) and 0.0041 at low resolution, at two orders of magnitude less extraction compute. Restricting the pass to the 149\,s window costs a further 0.0046, and the context lead recovers it.

\paragraph{MIRAGE head with LoRA.}
With LoRA on single-pass window-local features, the MIRAGE head reaches 0.3247 at a 60\,s lead, above the 0.3187 of the sliding-window protocol it was designed for. With an adapted backbone, the single-pass design matches or exceeds the sliding window. The head is trained in two stages, LoRA under the frozen head and then jointly with its per-subject output layer, so it receives twice the linear readout's 1{,}429 steps. At matched lead with both readouts re-fit, the linear readout is still ahead by $+0.0020$ [$+0.0008$, $+0.0033$] over four seed-matched pairs (0.3261 against 0.3241). Lead and readout largely do not interact: going from 30 to 60\,s is worth $+0.0010$ to the head and $+0.0012$ to the linear readout independently.

\paragraph{Context lead.}
Step time grows from 9.7\,s at no lead to 14.4\,s at 120\,s context lead. Accuracy is flat past 60\,s, so we use 60\,s. Each cell is a single run, so we compare cells with a paired parcel bootstrap. The first 30\,s of lead is worth $+0.00396$ and the next 30\,s $+0.00142$, both with intervals above zero and both well above the seed s.d.\ of this configuration. The linear readout encodes the span $[t_0-\ell,\,t_1]$ for a lead $\ell$, so whole-clip encoding is the $\ell \to t_0$ limit of the context-length sweep, and the table has no separate whole-clip rows for it. The flat context-length sweep performance shows that 60\,s suffices at the resolution of this metric.

\begin{figure}[t]
\centering
\includegraphics[width=\linewidth]{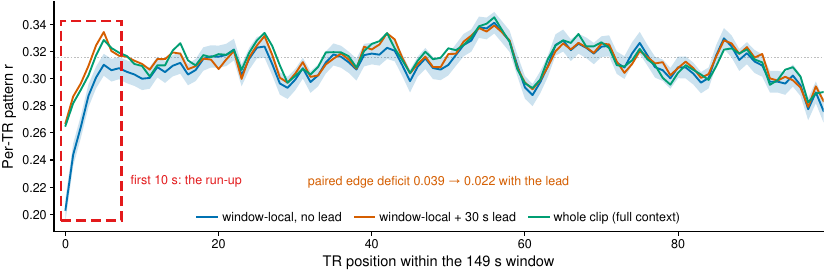}
\caption{\textbf{Accuracy within the window.} Accuracy against TR position inside the 149\,s window for three input designs, scored with the same frozen MIRAGE head on Friends s06 (606 complete interior windows; the band is the s.e.m.\ over windows for the no-lead configuration). The dashed box marks the first 10\,s, where the designs differ. Accuracy here is the per-TR spatial correlation between predicted and measured 1000-parcel patterns, which is defined at a single time point; it is not comparable to the per-parcel headline metric. Without a lead, a window starts 0.110 below its plateau and needs about five TRs to reach it. With a 30\,s lead it starts 0.048 below plateau, and with the whole clip 0.051. The first window of each clip is excluded because of a clip-onset transient.}
\label{fig:within-window}
\end{figure}

\paragraph{Within-window accuracy.}
Fig.~\ref{fig:within-window} shows the same effect per timestep. Accuracy is lower at the start of the window than in the middle, by $0.0393 \pm 0.0043$ with no lead, $0.0220 \pm 0.0042$ with a 30\,s lead and $0.0235 \pm 0.0044$ with the whole clip (paired differences). As with the overall score, the benefit of the lead saturates by 60\,s. Missing context therefore accounts for only 0.0173 of the start-of-window deficit; the remaining ${\sim}0.022$ persists even when the model sees the whole clip. Accuracy also falls at the end of the window. These drops are close to the residual start-of-window deficit, so stimulus beyond the window boundary is worth at most 0.003, within the s.e.m.\ (see also App.~\ref{app:prefix}).

\section{Efficiency and the multi-adapter kernel}
\label{app:efficiency}

\begin{figure}[t]
\centering
\includegraphics[width=\linewidth]{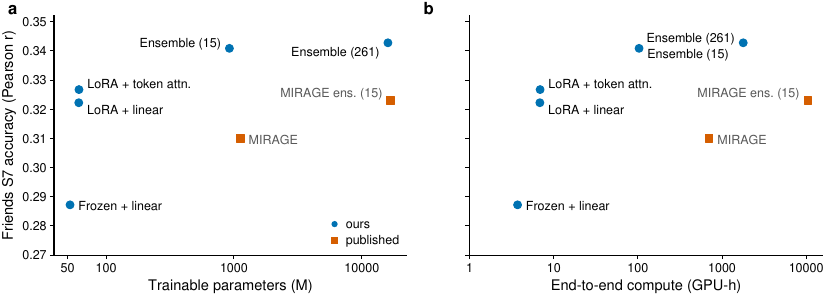}
\caption{\textbf{Accuracy against cost.} Held-out Friends S7 accuracy against trainable parameters \emph{(a)} and against end-to-end compute including feature extraction \emph{(b)}, both on log scales. Published baselines are squares and our models circles; an $N$-model ensemble is plotted at $N$ times a member's cost.}
\label{fig:efficiency}
\end{figure}

Fig.~\ref{fig:efficiency} places our models and the published baselines on two cost axes: trainable parameters, and end-to-end GPU-hours including feature extraction. Our single models use an order of magnitude fewer trainable parameters than the frozen-feature baselines and two orders of magnitude less compute than the sliding-window protocol, at equal or better accuracy.

\paragraph{Compute breakdown.}
Tab.~\ref{tab:compute} breaks down the cost of both configurations on one A800-80GB over the same 331-clip corpus.

\emph{Extraction.} The sliding-window protocol performs $2{,}758$ forward passes per clip, one per cutoff on the 2\,Hz grid and one per word onset, against $5.1$ for a single causal pass, i.e.\ $545\times$ fewer forwards. Our forward passes are longer (${\approx}14{,}900$ tokens each against a capped prefix), so the saving in GPU-hours is $352\times$: $700.6$ against $1.99$ GPU-h. We measured this on our own hardware.

\emph{End-to-end compute.} Our total comprises extraction ($1.99$), adapter training ($4.84$) and the closed-form re-fit ($1.67$), $8.5$ GPU-h against the baseline's $702.5$, an $83\times$ reduction. We include the re-fit because every accuracy we report is post-re-fit. Without it the total is $6.8$ GPU-h ($103\times$). The baseline total includes its head training ($1.9$ GPU-h, from our reproduction of the published configuration).

The sliding-window extraction was measured over the 49 validation clips and scaled to the full corpus; these clips are $1.1\%$ longer than the corpus mean, so this slightly overestimates the cost. The two protocols also compute different representations: the sliding window re-encodes a capped prefix (60\,s of audio, 4\,s of video, 1{,}024 words) at every cutoff, whereas a single pass with a 60\,s lead gives each timestep the whole window's joint context. The ratio compares the two protocols as run, not two ways of computing the same features.

\begin{table}[t]
\caption{\textbf{Compute for the two single-model configurations.} Both columns use one A800-80GB, the same 331-clip corpus and the same 336$\times$224 media cache (built once on CPU and shared). \emph{Sliding window} re-encodes a capped prefix (60\,s audio, 4\,s video, 1{,}024 words) at every cutoff; \emph{single causal pass} encodes each 149\,s window once with a 60\,s lead. Sliding-window extraction was measured over the 49 validation clips and scaled to 331 clips. Baseline training is our reproduction of the published head recipe. The closed-form re-fit is included in our column because all reported accuracies are post-re-fit. Extraction is 352$\times$ cheaper and the end-to-end cost 83$\times$ lower.}
\label{tab:compute}
\centering
\footnotesize
\setlength{\tabcolsep}{4pt}
\begin{tabular}{lrr}
\toprule
 & Frozen + 1{,}123M readout & LoRA + 52.3M readout \\
\midrule
\multicolumn{3}{l}{\emph{Hardware and precision}} \\
Device & \multicolumn{2}{c}{1$\times$ A800-80GB} \\
Precision & \multicolumn{2}{c}{bf16 backbone, fp32 patch-embed conv} \\
\midrule
\multicolumn{3}{l}{\emph{Stimulus encoding}} \\
Protocol & sliding window & single causal pass \\
Forward passes / clip & 2{,}758 & 5.1 \\
Tokens / forward & ${\leq}2{,}000$ & 14{,}890 \\
Seconds / forward & 2.76 & 4.28 \\
Extraction, 331 clips (GPU-h) & \textbf{701} & \textbf{1.99} \\
\midrule
\multicolumn{3}{l}{\emph{Training}} \\
Reads & stored features & live re-encode from media \\
Batch & 16 windows & 1 window (400 target rows) \\
Optimizer steps & 5{,}670 (15 epochs) & 1{,}429 (1 epoch) \\
Backward through backbone & no & yes, 32 of 48 layers \\
Seconds / step & -- & 12.0 \\
Training (GPU-h) & 1.88 & 4.84 \\
Closed-form re-fit (GPU-h) & -- & 1.67 \\
\midrule
Trainable parameters & 1{,}123M & 61.2M \\
\textbf{End-to-end (GPU-h)} & \textbf{702} & \textbf{8.5} \\
\bottomrule
\end{tabular}
\end{table}

\begin{figure}[t]
\centering
\includegraphics[width=\linewidth]{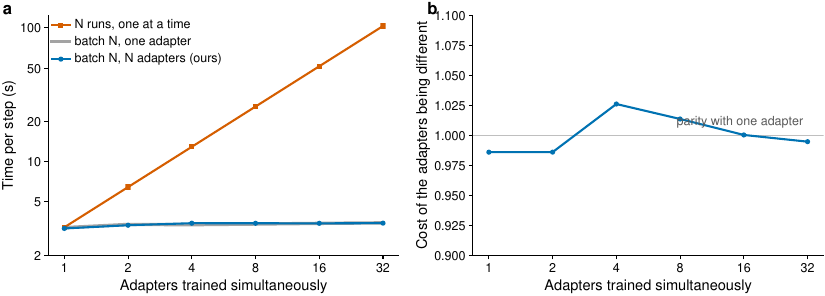}
\caption{\textbf{Multi-adapter kernel benchmark.}
\texttt{Qwen3-Omni} truncated at the feature extraction layer, rank-16 adapters on $q/k/v/o$, sequence length 128, one sample per adapter, one A800, median of 5 timed steps after 2 warm-up steps. 
\emph{(a)} One fused batched gather-matmul over $N$ different adapters (blue) matches the same batch through one adapter (grey); running the $N$ jobs one at a time (orange) grows linearly in $N$.
\emph{(b)} The ratio of the fused pass to the single-adapter batch is $1.00\times$ throughout, up to run-to-run noise of $\pm0.03$.}
\label{fig:kernel}
\end{figure}

\paragraph{Multi-adapter kernel.}
Adapters are small and structurally identical, so the routed low-rank update for a whole pool can be written as one batched gather-matmul instead of $N$ separate ones. Fig.~\ref{fig:kernel} benchmarks our Triton kernel for this on the backbone of \S\ref{sec:method} (the data behind
Fig.~\ref{fig:scaleout}a). Against $N$ runs taken one at a time, the fused pass is $29.8\times$ faster at $N=32$. The speedup results from GPU utilization, not a saving in actual compute in FLOPs: a batch-1 step on a sparse mixture-of-experts backbone leaves the GPU underused, and the fused step at $N=32$ costs only $1.08\times$ a single-adapter step. Against the same batch of $N$ through a single adapter, the fused pass costs $1.00\times$ ($0.99$--$1.03$ over $N \le 32$), so using different adapters adds no cost over batching. Memory is the limit: at rank 16 the pool and its optimizer state fill an 80GB card between 32 and 64 adapters.

\section{Selection and evaluation protocol}
\label{app:protocol}

\paragraph{Selection on validation set.}
Every design choice was made on the Friends season-6 validation split, including extraction design, context lead, feature extraction layer, LoRA rank, learning rates, readout family, ensemble membership, and combiner weights. Every model was finalized using season 6 before submission, and held-out leaderboard scores were only reported, never used to select, rank, or tune a model. Our submissions were made to the public leaderboard in the post-challenge phase of Algonauts 2025.

\paragraph{Feature extraction layer.}
The readout reads decoder layer 32 of 48, and the decoder is truncated there. The feature extraction layer was chosen on a layer comparison using ridge readouts on frozen features and checked against trained runs at the neighboring layers (Tab.~\ref{tab:hp-tap}, App.~\ref{app:hparams}).

\paragraph{Seed variability.}
Each of our rows in Tab.~\ref{tab:main} is a single run. Across seeds, LoRA\,+\,linear scores $0.32609 \pm 0.00024$ (15 seeds) and LoRA\,+\,cross-attn $0.33070 \pm 0.00023$ (3 seeds). Rows that differ by less than about 0.0008 are therefore not separated by the table alone, and the comparisons in \S\ref{sec:main} use a paired bootstrap or seed statistics.

\paragraph{Ridge penalties.}
The closed-form re-fit accumulates its moments on the training windows only. Per-parcel penalties are chosen from a fixed five-point grid on a deterministic strided 15\% holdout of the training windows (128{,}400 fit rows and 22{,}800 holdout rows per subject). The chosen penalty is then re-solved on fit and holdout together. Season 6 is never used to fit or select a penalty; it is only scored, over its 250 windows.

\paragraph{Ensemble stacking.}
The per-parcel minimum-variance combiner \citep{perrone1993disagree} is fitted on season 6 out of fold, split by episode. Windows within an episode share stimulus and scanner state, so a window-level split would leak across folds. Rows in one half are predicted by weights fitted on the other, and one scoring pass covers every row exactly once. For an unfitted rule such as the uniform mean, out-of-fold and in-sample scores coincide by construction, which serves as a check on the procedure. The ensemble rows of Tab.~\ref{tab:main} are the out-of-fold stack. For the 261-model pool the uniform mean scores 0.34007 and the stack 0.34834.

\paragraph{Subset search.}
The 15-model subset is chosen by greedy selection with one-swap refinement under a nested protocol: the subset and its weights are both chosen on one episode half and scored on the other, and the two quarters within each half set greedy's stopping step. The assembled out-of-fold predictions again cover every row once. The procedure scores 0.34651, and the selection bias of the reported roster is $+0.00035$.

\section{Hyperparameter selection}
\label{app:hparams}

\begin{table}[t]
\caption{\textbf{LoRA rank.} Shipped recipe otherwise unchanged
(\textsc{Qwen3-Omni}, linear readout, extraction layer 32, 60\,s lead, LoRA lr
$6\times10^{-4}$, 1{,}429 steps), Friends s06 subject-mean after the
closed-form re-fit. Rank and readout learning rate were not varied
independently; the matched comparison is the two rows at
$3\times10^{-4}$. Diverged runs are counted in parentheses.}
\label{tab:hp-rank}
\centering
\footnotesize
\begin{tabular}{rrrrr}
\toprule
Rank & Readout lr & Runs & s06 (mean) & s.d. \\
\midrule
16 & 3$\times10^{-4}$ & 4 & 0.3262 & 0.00007 \\
32 & 3$\times10^{-4}$ & 1 & 0.3268 & -- \\
32 & 5$\times10^{-4}$ & 9 & 0.3257 & 0.00109 \\
64 & 5$\times10^{-4}$ & 4 (+1 diverged) & 0.3237 & 0.00128 \\
128 & 5$\times10^{-4}$ & 0 (+2 diverged) & -- & -- \\
\bottomrule
\end{tabular}
\end{table}

\begin{table}[t]
\caption{\textbf{Feature-extraction layer.} \emph{Frozen} is the closed-form ridge readout
on unadapted features (no adapter, no training); \emph{trained} is the full
recipe, one seed per layer, after the re-fit. The frozen screen peaks at
layer 30; the trained runs at layers
31--34 lie within 0.0006 of one another, inside the
0.00077 seed spread of this configuration.}
\label{tab:hp-tap}
\centering
\footnotesize
\begin{tabular}{rrr}
\toprule
Feature-extraction layer & Frozen & Trained \\
\midrule
16 & 0.2682 & -- \\
24 & 0.2839 & -- \\
30 & 0.2889 & 0.3252 \\
31 & 0.2885 & 0.3260 \\
32 & 0.2881 & 0.3263 \\
33 & 0.2872 & 0.3258 \\
34 & 0.2876 & 0.3264 \\
40 & 0.2836 & -- \\
44 & 0.2784 & -- \\
\bottomrule
\end{tabular}
\end{table}

\begin{table}[t]
\caption{\textbf{Learning rates.} Measured in an earlier configuration
without context lead or closed-form re-fit, so these are pre-re-fit
validation scores about 0.013 below the final recipe, comparable to each
other but not to Tab.~\ref{tab:main}. The per-modality projection uses
$2\times$ the readout rate. The first row freezes the adapter (readout-only
control).}
\label{tab:hp-lr}
\centering
\footnotesize
\begin{tabular}{rrr}
\toprule
LoRA lr & Readout lr & s06 (pre-re-fit) \\
\midrule
0 & 2e-5 & 0.2818 \\
1e-3 & 1e-4 & 0.3071 \\
1e-3 & 3e-4 & 0.3131 \\
6e-4 & 1e-3 & 0.3076 \\
6e-4 & 3e-4 & 0.3134 \\
\bottomrule
\end{tabular}
\end{table}

\begin{table}[t]
\caption{\textbf{A second epoch overfits.} Validation score through a
2858-step (two-epoch) run, same configuration as
Tab.~\ref{tab:hp-lr}. The peak is at step 1400, the end of the
first pass over the 1{,}429 training windows. Matched one- versus two-epoch
pairs give $-0.0147$ at readout lr $3\times10^{-4}$, $-0.0047$ at readout lr $1\times10^{-4}$.}
\label{tab:hp-epoch}
\centering
\scriptsize
\setlength{\tabcolsep}{3pt}
\begin{tabular}{lrrrrrrrrrrrrrrr}
\toprule
Step & 200 & 400 & 600 & 800 & 1000 & 1200 & 1400 & 1600 & 1800 & 2000 & 2200 & 2400 & 2600 & 2800 & 2858 \\
\midrule
s06 & 0.2789 & 0.2856 & 0.2877 & 0.2921 & 0.2977 & 0.3007 & \textbf{0.3052} & 0.3009 & 0.2999 & 0.2996 & 0.3017 & 0.3019 & 0.3031 & 0.3030 & 0.3028 \\
\bottomrule
\end{tabular}
\end{table}

All values below were chosen on Friends season 6 (App.~\ref{app:protocol}). We report the full selection process below.

\paragraph{LoRA rank (Tab.~\ref{tab:hp-rank}).}
We hold $\alpha = 2r$ fixed, so raising the rank also raises the effective scale of the update at a fixed LoRA learning rate; rank and learning rate are therefore coupled. At the one matched setting (readout lr $3\times10^{-4}$), rank 32 is $+0.0006$ above rank 16, inside the 0.00077 seed spread of this configuration. The rank-32 configuration also varies more across seeds (0.00109 over nine seeds, against 0.00007 over four rank-16 runs). Rank 64 loses $0.0025$ and one of its five seeds diverged; both rank-128 seeds diverged. We use rank 16, the smallest rank that no other rank beats beyond seed noise. Higher ranks were run at the learning rate tuned for rank 16 and may train better at a lower rate; on \textsc{Ming-flash-omni}, for instance, rank 48 diverges at this rate and
trains at $3\times10^{-4}$.

\paragraph{Feature extraction layer (Tab.~\ref{tab:hp-tap}).}
A closed-form ridge readout on unadapted features peaks at layer 30, and the full configuration peaks at layer 34. The trained runs at layers 31--34 lie within 0.0006 of one another, inside the 0.00077 seed spread, so layer 32 is a good feature extraction layer within a flat neighborhood rather than a verified optimum. The layer tuning experiment on frozen features locates the neighborhood: it falls to 0.2682 at layer 16 and 0.2784 at layer 44. 

\paragraph{Schedule length (Tab.~\ref{tab:hp-epoch}).}
We train for exactly one epoch. A two-epoch run peaks at step 1{,}400, the end of the first pass over the 1{,}429 training windows, and stays below that peak throughout the second epoch. Two matched one- versus two-epoch pairs give $-0.0147$ and $-0.0047$. The training set already contains all of Friends seasons 1--5 and \emph{movie10} (1{,}429 windows after removing windows that extend past their media or lack fMRI). 

\paragraph{Learning rates (Tab.~\ref{tab:hp-lr}).}
The per-modality projection is held at twice the readout rate, so the grid is two-dimensional. The readout rate has an interior optimum at $3\times10^{-4}$, which beats $1\times10^{-4}$ by 0.0060 at LoRA lr $1\times10^{-3}$ and $1\times10^{-3}$ by 0.0058 at LoRA lr $6\times10^{-4}$. The LoRA rate is flat between $6\times10^{-4}$ and $1\times10^{-3}$ at this readout rate (0.3134 against 0.3131, with a two-seed spread of 0.0004); we use $6\times10^{-4}$. Freezing the adapter and training only the readout gives 0.2818. These runs predate the context lead and the closed-form re-fit, so they are pre-re-fit scores about 0.013 below the final configuration and are comparable only to one another. The rates were not re-tuned afterwards.

\paragraph{Interaction with the re-fit.}
The readout-rate effect (0.0058) is close to the value of the closed-form re-fit ($+0.0061$, \S\ref{sec:ablation}). Part of what the readout rate controls is how far $W$ drifts from the ridge optimum during training, and the re-fit removes that drift. After the re-fit, the same readout-rate comparison is $+0.00025$, and the rank-32 advantage shrinks in the same way. This explains why the re-fit’s gain is comparable to that of the readout rate (\S\ref{sec:ablation}). 

\section{The effect of local non-causality in the backbone}
\label{app:prefix}

Single-pass extraction assumes that a state at time $t$ depends only on the stimulus up to $t$. We encode a span once, then re-encode strict prefixes of the same span from the same start, so the shared tokens and their position ids are identical. If causal truncation held exactly, feature extraction layer states at the shared timesteps would agree.

As a control, the same span encoded twice agrees at cosine 1.0000 (minimum 0.99999) on all three streams, so the pipeline is deterministic. Cuts are placed on chunk boundaries, and the final chunk before each cut is excluded, because a truncated span ends with its own closing tokens where the longer span continues with content.

\begin{table}[ht]
\centering
\footnotesize
\setlength{\tabcolsep}{5pt}
\caption{Cosine between representations at shared timesteps, prefix pass
against single pass, on \emph{friends\_s01e01a} with a 209\,s span, feature extraction layer 32.}
\label{tab:prefix}
\begin{tabular}{lrrr}
\toprule
Prefix ends at & audio & vision & text \\
\midrule
140\,s & 0.9579 & 0.9895 & 0.9960 \\
160\,s & 0.9569 & 0.9908 & 0.9970 \\
180\,s & 0.9553 & 0.9901 & 0.9969 \\
200\,s & 0.9584 & 0.9906 & 0.9973 \\
\midrule
same span twice & 1.0000 & 1.0000 & 1.0000 \\
\bottomrule
\end{tabular}
\end{table}

The states differ by more than the control allows (Tab.~\ref{tab:prefix}). The location of the deviation shows that it is not future stimulus leaking backwards. The decoder is causally masked, the audio encoder attends within 8\,s blocks, and the vision encoder uses 1\,s blocks. A leak would therefore be confined to the chunk next to the cut, which is excluded. Instead, the \emph{first} chunk deviates most: per 20\,s chunk of the 140\,s prefix, audio cosine is 0.919, 0.980, 0.979, 0.969, 0.958 and 0.942, from the start of the span onwards. The first chunk lies 120\,s before the cut, where both passes see identical tokens. We attribute the deviation to length-dependent bf16 numerics: attention kernels tile different sequence lengths differently, and top-$k$ expert routing can turn a rounding difference into a different expert.

A feature at $t$ can thus depend on stimulus after $t$ only up to the end of its 8\,s audio block. The two extraction protocols differ mainly in which prefix they give the model, which is what the context-lead length sweep experiment measures. At matched media, replacing the sliding window with one whole-clip pass costs 0.0038 under the frozen MIRAGE head (Tab.~\ref{tab:input-design}), and the context lead recovers more than that. The context length sweep itself compares prefixes under one numerical regime, since train and evaluation spans have the same length throughout.

\section{Ensemble configurations}
\label{app:roster}

The $y$ axis of Fig.~\ref{fig:scaleout}b is $-\log_{10}(m_\infty - r)$. Equal steps are equal fractions of the remaining gap, and $m(n) = m_\infty - An^{-b}$ is a line of slope $b$. A straight line therefore describes the rate of convergence, not unbounded growth. Here $m_\infty$ is the limit of the \emph{mean} subset. Subsets scoring above it lie at $-\infty$ on this axis and are not shown; this happens from $n = 3$ on, and for 16\% of subsets at $n = 20$. At $n = 20$ the 5th--95th percentile range is 0.3360--0.3409 around a mean of 0.3388.

Tab.~\ref{tab:roster} lists every model in the 261-member pool with the configuration that distinguishes it. All members share the training configuration of \S\ref{sec:method} unless a column says otherwise.

{\footnotesize
\setlength{\tabcolsep}{4pt}
\setlength{\LTcapwidth}{\linewidth}
\begin{longtable}{@{}>{\raggedright\arraybackslash\hangindent=1.2em\hangafter=1}p{\dimexpr\linewidth-3.0cm-4\tabcolsep\relax}>{\raggedleft\arraybackslash}p{1.5cm}>{\raggedleft\arraybackslash}p{1.5cm}@{}}
\caption{\textbf{The 261-model pool.} Each row is a distinct configuration; members differing only in random seed are grouped, with the seed count given. Entries list only where a member departs from the shipped recipe (\texttt{Qwen3-Omni} backbone, linear readout, feature extraction layer 32, rank 16, 60\,s lead, 1{,}429 steps). Counts sum to 261.}
\label{tab:roster}\\
\toprule
Configuration & Members & Seeds \\
\midrule
\endfirsthead
\caption[]{\emph{The 261-model pool, continued.}}\\
\toprule
Configuration & Members & Seeds \\
\midrule
\endhead
\midrule
\multicolumn{3}{r}{\emph{continued on the next page}}\\
\endfoot
\midrule
Total & 261 & \\
\bottomrule
\endlastfoot
shipped recipe & 34 & 15 seeds \\
rank 32, readout lr 0.0005 & 27 & 13 seeds \\
lead 30 & 16 & 8 seeds \\
backbone \texttt{Qwen3.8-27B}, chunk 2, no audio & 10 & 6 seeds \\
readout cross-attn (tokens), tokens stream & 8 & 3 seeds \\
bag 0.6 & 6 & 2 seeds \\
readout attn\_pq & 6 & 4 seeds \\
backbone \texttt{Gemma 3 27B}, feature extraction layer 35, chunk 2, no audio & 5 & 4 seeds \\
backbone \texttt{Qwen3.8-27B}, feature extraction layer 36, chunk 2, no audio & 5 & 5 seeds \\
rank 64, readout lr 0.0005 & 4 & 4 seeds \\
rank 8, lead 30 & 4 & 2 seeds \\
readout cross-attn (tokens), tokens raw & 4 & 3 seeds \\
backbone \texttt{Ming-flash 104B}, readout cross-attn (tokens), feature extraction layer 24, rank 48, LoRA lr 0.0003, scope attn\_shared, tokens stream & 3 & 3 seeds \\
backbone \texttt{Qwen3.8-27B}, rank 32, chunk 2, no audio & 3 & 2 seeds \\
bag 0.4 & 3 & 1 seed \\
LoRA lr 2e-05, readout lr 1e-05, steps 715 & 2 & 2 seeds \\
backbone \texttt{Gemma 3 27B}, readout mlp, feature extraction layer 35, chunk 2, no audio & 2 & 2 seeds \\
backbone \texttt{LCO-Omni 7B} & 2 & 1 seed \\
backbone \texttt{LCO-Omni 7B}, steps 1 & 2 & 1 seed \\
backbone \texttt{Ming-flash 104B}, feature extraction layer 20, scope attn\_shared & 2 & 1 seed \\
backbone \texttt{Ming-flash 104B}, feature extraction layer 24, rank 48, LoRA lr 0.0003, scope attn\_shared & 2 & 2 seeds \\
backbone \texttt{Ming-flash 104B}, feature extraction layer 24, rank 48, LoRA lr 0.0003, scope attn\_shared\_router & 2 & 1 seed \\
backbone \texttt{MiniCPM-o} & 2 & 2 seeds \\
backbone \texttt{MiniCPM-o}, steps 1 & 2 & 1 seed \\
backbone \texttt{Nemotron Omni} & 2 & 2 seeds \\
backbone \texttt{Qwen2.5-Omni 7B} & 2 & 1 seed \\
backbone \texttt{Qwen2.5-Omni 7B}, steps 1 & 2 & 1 seed \\
backbone \texttt{Qwen3.8-27B}, no audio & 2 & 2 seeds \\
backbone \texttt{Qwen3.8-27B}, feature extraction layer 40, chunk 2, no audio & 2 & 2 seeds \\
backbone \texttt{Qwen3.8-27B}, feature extraction layer 44, chunk 2, no audio & 2 & 2 seeds \\
backbone \texttt{Qwen3.8-27B}, feature extraction layer 48, chunk 2, no audio & 2 & 2 seeds \\
backbone \texttt{Qwen3.8-27B}, feature extraction layer 52, chunk 2, no audio & 2 & 2 seeds \\
backbone \texttt{Qwen3.8-27B}, feature extraction layer 56, chunk 2, no audio & 2 & 2 seeds \\
backbone \texttt{Qwen3.8-27B}, feature extraction layer 60, chunk 2, no audio & 2 & 2 seeds \\
backbone \texttt{Qwen3.8-27B}, feature extraction layer 64, chunk 2, no audio & 2 & 2 seeds \\
rank 32, lead 90, readout lr 0.0005 & 2 & 2 seeds \\
rank 32, readout lr 0.0005, chunk 2 & 2 & 2 seeds \\
readout cross-attn (layers) & 2 & 1 seed \\
readout cross-attn (tokens), tokens stream\_sp2x1 & 2 & 2 seeds \\
readout mlp & 2 & 1 seed \\
steps 1466 & 2 & 2 seeds \\
LoRA lr 0, readout lr 0, steps 1 & 1 & 1 seed \\
LoRA lr 0.0001, readout lr 5e-05, steps 715 & 1 & 1 seed \\
LoRA lr 0.0002, readout lr 0.0001, steps 715 & 1 & 1 seed \\
LoRA lr 5e-05, readout lr 2.5e-05, steps 715 & 1 & 1 seed \\
backbone \texttt{Gemma 3 27B}, feature extraction layer 30, chunk 2, no audio & 1 & 1 seed \\
backbone \texttt{Gemma 3n E4B}, readout cross-attn (tokens), feature extraction layer 25, tokens stream & 1 & 1 seed \\
backbone \texttt{Gemma 3n E4B}, readout cross-attn (tokens), feature extraction layer 30, tokens stream & 1 & 1 seed \\
backbone \texttt{Gemma 3n E4B}, feature extraction layer 25 & 1 & 1 seed \\
backbone \texttt{Gemma 3n E4B}, feature extraction layer 30 & 1 & 1 seed \\
backbone \texttt{LCO-Omni 7B}, readout cross-attn (tokens), feature extraction layer 26, tokens stream & 1 & 1 seed \\
backbone \texttt{LCO-Omni 7B}, feature extraction layer 20 & 1 & 1 seed \\
backbone \texttt{LCO-Omni 7B}, feature extraction layer 26 & 1 & 1 seed \\
backbone \texttt{LCO-Omni 7B}, feature extraction layer 28 & 1 & 1 seed \\
backbone \texttt{Ming-flash 104B} & 1 & 1 seed \\
backbone \texttt{Ming-flash 104B}, readout attn\_branch, feature extraction layer 24, rank 48, LoRA lr 0.0003, scope attn\_shared & 1 & 1 seed \\
backbone \texttt{Ming-flash 104B}, readout cross-attn (tokens), feature extraction layer 16, tokens grid & 1 & 1 seed \\
backbone \texttt{Ming-flash 104B}, readout cross-attn (tokens), feature extraction layer 16, tokens raw & 1 & 1 seed \\
backbone \texttt{Ming-flash 104B}, readout cross-attn (tokens), feature extraction layer 16, tokens stream & 1 & 1 seed \\
backbone \texttt{Ming-flash 104B}, readout cross-attn (tokens), feature extraction layer 20, tokens stream & 1 & 1 seed \\
backbone \texttt{Ming-flash 104B}, readout cross-attn (tokens), feature extraction layer 24, rank 48, LoRA lr 0.0003, scope attn\_shared, tokens grid & 1 & 1 seed \\
backbone \texttt{Ming-flash 104B}, readout cross-attn (tokens), feature extraction layer 24, tokens stream & 1 & 1 seed \\
backbone \texttt{Ming-flash 104B}, feature extraction layer 16 & 1 & 1 seed \\
backbone \texttt{Ming-flash 104B}, feature extraction layer 16, steps 2858 & 1 & 1 seed \\
backbone \texttt{Ming-flash 104B}, feature extraction layer 20 & 1 & 1 seed \\
backbone \texttt{Ming-flash 104B}, feature extraction layer 20, LoRA lr 0.0002, scope attn\_shared & 1 & 1 seed \\
backbone \texttt{Ming-flash 104B}, feature extraction layer 20, rank 48, LoRA lr 0.0002, scope attn\_shared & 1 & 1 seed \\
backbone \texttt{Ming-flash 104B}, feature extraction layer 20, rank 48, LoRA lr 0.0003, scope attn\_shared & 1 & 1 seed \\
backbone \texttt{Ming-flash 104B}, feature extraction layer 20, rank 48, LoRA lr 0.00045, scope attn\_shared & 1 & 1 seed \\
backbone \texttt{Ming-flash 104B}, feature extraction layer 24 & 1 & 1 seed \\
backbone \texttt{Ming-flash 104B}, feature extraction layer 24, scope attn\_shared & 1 & 1 seed \\
backbone \texttt{Ming-flash 104B}, feature extraction layer 28 & 1 & 1 seed \\
backbone \texttt{MiniCPM-o}, readout attn\_branch, feature extraction layer 28 & 1 & 1 seed \\
backbone \texttt{MiniCPM-o}, readout cross-attn (tokens), feature extraction layer 28, tokens stream & 1 & 1 seed \\
backbone \texttt{MiniCPM-o}, feature extraction layer 28 & 1 & 1 seed \\
backbone \texttt{Nemotron Omni}, readout attn\_branch, feature extraction layer 44 & 1 & 1 seed \\
backbone \texttt{Nemotron Omni}, readout cross-attn (tokens), feature extraction layer 44, tokens stream & 1 & 1 seed \\
backbone \texttt{Nemotron Omni}, steps 1 & 1 & 1 seed \\
backbone \texttt{Nemotron Omni}, feature extraction layer 44 & 1 & 1 seed \\
backbone \texttt{Phi-4-MM}, readout attn\_branch, feature extraction layer 20 & 1 & 1 seed \\
backbone \texttt{Phi-4-MM}, readout cross-attn (tokens), feature extraction layer 20, tokens stream & 1 & 1 seed \\
backbone \texttt{Phi-4-MM}, feature extraction layer 20 & 1 & 1 seed \\
backbone \texttt{Qwen2.5-Omni 3B}, readout attn\_branch, feature extraction layer 30 & 1 & 1 seed \\
backbone \texttt{Qwen2.5-Omni 3B}, readout cross-attn (tokens), feature extraction layer 30, tokens stream & 1 & 1 seed \\
backbone \texttt{Qwen2.5-Omni 3B}, feature extraction layer 30 & 1 & 1 seed \\
backbone \texttt{Qwen2.5-Omni 7B}, readout attn\_branch, feature extraction layer 24 & 1 & 1 seed \\
backbone \texttt{Qwen2.5-Omni 7B}, readout cross-attn (tokens), feature extraction layer 24, tokens stream & 1 & 1 seed \\
backbone \texttt{Qwen2.5-Omni 7B}, feature extraction layer 24 & 1 & 1 seed \\
backbone \texttt{Qwen3.8-27B}, LoRA lr 0, readout lr 0, chunk 2, no audio, steps 1 & 1 & 1 seed \\
backbone \texttt{Qwen3.8-27B}, LoRA lr 0.001, chunk 2, no audio & 1 & 1 seed \\
backbone \texttt{Qwen3.8-27B}, chunk 2, no audio, scope attn & 1 & 1 seed \\
backbone \texttt{Qwen3.8-27B}, chunk 2, no audio, steps 1 & 1 & 1 seed \\
bag 0.8 & 1 & 1 seed \\
chunk 10 & 1 & 1 seed \\
chunk 2 & 1 & 1 seed \\
chunk 2, no audio & 1 & 1 seed \\
chunk 5 & 1 & 1 seed \\
lead 120 & 1 & 1 seed \\
lead 45 & 1 & 1 seed \\
lead 75 & 1 & 1 seed \\
lead 90 & 1 & 1 seed \\
no audio & 1 & 1 seed \\
rank 32 & 1 & 1 seed \\
rank 32, lead 120 & 1 & 1 seed \\
rank 32, lead 75, readout lr 0.0005 & 1 & 1 seed \\
readout cross-attn (tokens), tokens grid & 1 & 1 seed \\
readout lr 0.0005 & 1 & 1 seed \\
feature extraction layer 30 & 1 & 1 seed \\
feature extraction layer 31 & 1 & 1 seed \\
feature extraction layer 33 & 1 & 1 seed \\
feature extraction layer 34 & 1 & 1 seed \\
feature extraction layer 34, rank 32, readout lr 0.0005 & 1 & 1 seed \\
\end{longtable}
}

\section{Alternative PEFT methods}
\label{app:peft}

All other adaptation results in this paper use LoRA. To test whether the gain depends on the method, we ran two further PEFT families under the same protocol: same backbone, feature extraction layer, context lead, readout, readout learning rate, step budget and seed. Only the adaptation method changes, and all configurations are compared with the same frozen control.

\paragraph{Methods.}
IA$^3$ \citep{ia3} learns three element-wise rescaling vectors per decoder block, on the keys, the values and the FFN intermediate activations, and adds no weights. We run the standard attention + FFN scope and a variant that also rescales the intermediate activations of the MoE experts, which LoRA does not touch. Bottleneck adapters \citep{houlsby2019parameter} insert $x \mapsto x + W_{\mathrm{up}}\,\mathrm{gelu}(W_{\mathrm{down}}x)$ into the residual stream. Unlike LoRA, this update is nonlinear and cannot be merged into the frozen weights. We run the Houlsby placement (after both the attention sub-layer and the FFN) and the FFN-only placement of \citet{pfeiffer2021adapterfusion}, both at rank $32$. At this rank the Houlsby adapter has $8{,}521{,}728$ parameters against LoRA's $8{,}912{,}896$, so the two are budget-matched.

\begin{table}[t]
\caption{\textbf{Three PEFT families at one protocol.} Friends s06 subject-mean after the closed-form re-fit, at the main configuration (\textsc{Qwen3-Omni}, linear readout, feature extraction layer 32, 60\,s lead, 1{,}429 steps, one seed), with only the adaptation method changed. The column \emph{Params} is backbone-side trainable parameters, read from each checkpoint; the 52.3M readout is common to every row and is excluded. The column \emph{of LoRA} is the gain over the frozen backbone as a fraction of LoRA's, and the last column is that gain per million adapter parameters. IA$^3$ learns three element-wise rescaling vectors per block (keys, values, FFN intermediate) and adds no weights; a bottleneck adapter inserts $x + W_{\mathrm{up}}\,\mathrm{gelu}(W_{\mathrm{down}}x)$ after the attention sub-layer and the FFN (Houlsby) or after the FFN alone (Pfeiffer). Each family runs at its own screened rate ($1\times10^{-2}$, $3\times10^{-4}$, $6\times10^{-4}$), since a PEFT rate does not transfer between families.}
\label{tab:peft}
\centering
\footnotesize
\setlength{\tabcolsep}{4pt}
\begin{tabular}{lrrrrr}
\toprule
Method & Params & s06 & $\Delta$ frozen & \% of LoRA gain & $\Delta$ / M params \\
\midrule
Frozen backbone (readout only) & 0 & 0.2907 & --- & --- & --- \\
IA$^3$, attention + FFN & 0.057M & 0.3127 & $+0.0220$ & 62\% & 0.3833 \\
IA$^3$, + MoE experts & 3.18M & 0.3177 & $+0.0270$ & 76\% & 0.0085 \\
Bottleneck adapter, FFN only ($r{=}32$) & 4.26M & 0.3164 & $+0.0257$ & 72\% & 0.0060 \\
Bottleneck adapter, attention + FFN ($r{=}32$) & 8.52M & 0.3190 & $+0.0283$ & 79\% & 0.0033 \\
LoRA $r{=}16$ (ours) & 8.91M & 0.3263 & $+0.0356$ & 100\% & 0.0040 \\
\bottomrule
\end{tabular}
\end{table}

\paragraph{Gain over the frozen backbone.}
All four non-LoRA configurations beat the frozen backbone by margins tens of times the $0.00077$ seed spread of this configuration. Each improves $998$ of $1{,}000$ parcels (pooled over subjects, paired block bootstrap against the frozen control) and every subject individually ($+0.021$ to $+0.031$). The benefit of adaptation is therefore not specific to LoRA, although LoRA gives the largest gain.

\paragraph{Budget-matched comparison.}
At $96\%$ of LoRA's parameters, the Houlsby adapter recovers $79\%$ of its gain, a deficit of $0.0073$, about ten times the seed spread. At matched size, the nonlinear, non-mergeable bottleneck adapter does not outperform LoRA. The FFN-only placement is within $0.0026$ of the Houlsby placement at half the parameters, a difference its bootstrap does not resolve. This matches the redundancy between adapter sites reported by \citet{pfeiffer2021adapterfusion}.

\paragraph{Parameter count.}
Scores do not follow adapter size. The three middle configurations span $3.18$M to $8.52$M parameters and lie within $0.0026$ of one another, with overlapping bootstrap intervals, so we do not rank them. Going from the smallest non-LoRA adapter to the largest multiplies parameters by $149$ and adds $0.0063$, whereas LoRA exceeds the Houlsby adapter by $0.0073$ at the same size. Per parameter, IA$^3$ is the most efficient: its $57{,}344$ coefficients ($0.64\%$ of the LoRA adapter) recover $62\%$ of LoRA's gain, about $96\times$ the gain per parameter.

\paragraph{Learning rates.}
Each family runs at its own tuned rate: $1\times10^{-2}$ for IA$^3$, $3\times10^{-4}$ for the bottleneck adapter and $6\times10^{-4}$ for LoRA. Both non-LoRA rates are interior optima of their grids (IA$^3$: six values from $3\times10^{-4}$ to $1\times10^{-1}$; bottleneck: four values from $1\times10^{-4}$ to $3\times10^{-3}$). With LoRA's rate, IA$^3$ would sit near the bottom of its curve. A plausible reason is numerical precision: a multiplicative rescaling smaller than ${\sim}0.4\%$ cannot be represented in a bf16 activation, which pushes a multiplicative adapter to a high rate, while an additive update has no such floor. 

\paragraph{Step time.}
A smaller adapter does not train faster. The IA$^3$ model takes $13.1$\,s per step and the $149\times$ larger Houlsby adapter $11.4$\,s, since step time is dominated by the backbone's forward and backward passes. A small adapter saves optimizer state and storage, not training time.

\section{The backbone comparison}
\label{app:backbones}

\begin{table}[t]
\caption{\textbf{The nine omni-modal backbones.} Total parameters from each checkpoint's own index (\textsc{Qwen3-Omni} and \textsc{Ming} are mixture-of-experts, with 3B and 6B active per token); feature extraction layer is chosen per backbone. All arms share the recipe of \S\ref{sec:main}: rank-16 LoRA, 60\,s context lead, closed-form re-fit, Friends s06 subject-mean.}
\label{tab:backbones}
\centering
\footnotesize
\begin{tabular}{lrrrrrr}
\toprule
Backbone & Params & Extraction layer & Frozen & +LoRA & $\Delta$ & +cross-attn \\
\midrule
Phi-4-MM & 4.7B & 20 & 0.2521 & 0.2841 & +0.0320 & 0.2888 \\
Qwen2.5-Omni 3B & 6.0B & 30 & 0.2807 & 0.3139 & +0.0331 & 0.3175 \\
Gemma 3n E4B & 7.8B & 25 & 0.2631 & 0.3026 & +0.0395 & 0.3049 \\
LCO-Omni 7B & 8.9B & 26 & 0.2683 & 0.3199 & +0.0517 & 0.3200 \\
MiniCPM-o & 9.4B & 28 & 0.2594 & 0.3249 & +0.0655 & 0.3265 \\
Qwen2.5-Omni 7B & 11.2B & 24 & 0.2748 & 0.3178 & +0.0431 & 0.3214 \\
Qwen3-Omni 30B & 31.7B & 32 & 0.2907 & 0.3263 & +0.0356 & 0.3310 \\
Nemotron Omni & 33.0B & 44 & 0.2780 & 0.3227 & +0.0447 & 0.3257 \\
Ming-flash 104B & 104.3B & 24 & 0.2856 & 0.3227 & +0.0371 & 0.3253 \\
\bottomrule
\end{tabular}
\end{table}

Tab.~\ref{tab:backbones} lists the configurations behind Fig.~\ref{fig:backbone}. Each configuration uses the configuration of \S\ref{sec:main} with only the backbone changed, at that backbone's own feature extraction layer, chosen from a frozen feature extraction layer sweep. The frozen row is the same run with the adapter's learning rate set to zero. Its readout and per-modality projection train at the configuration's own rates for the same 1{,}429 steps, followed by the same closed-form re-fit, so the frozen and adapted rows differ only in the adapter. Each backbone contributes one fixed run covering all four subjects, rather than its best-scoring run, because the number of runs differs widely across backbones.

\textsc{Qwen3.8-27B} and \textsc{Gemma-3-27B} appear only in
Tab.~\ref{tab:ablation}. They have no audio inputs, so their two-modality scores are not comparable here: removing audio alone costs $0.0232$ on \textsc{Qwen3-Omni} (App.~\ref{app:modality}).

\section{Why seeds are not a source of diversity}
\label{app:seedgeom}

\begin{figure}[t]
\centering
\includegraphics[width=\linewidth]{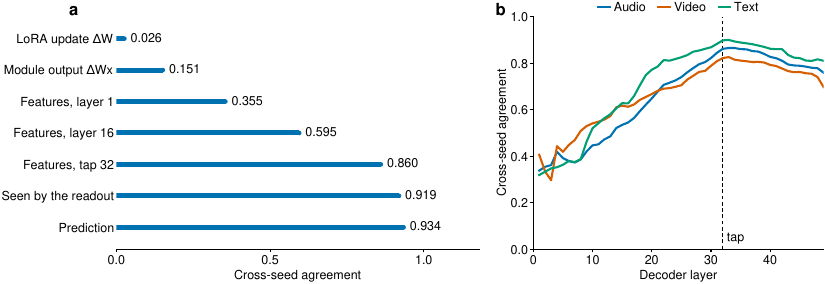}
\vspace{-20pt}
\caption{\textbf{Cross-seed agreement.} Mean agreement between adapters over the 15 seeds of the main configuration (105 pairs). 
\emph{(a)} At successive levels of the model, from near zero for the weights to near one for the prediction. Levels below the first are cosines between the two seeds' \emph{changes} relative to the frozen model, averaged over the audio, video, and text streams. 
\emph{(b)} Agreement at every decoder layer, per stream. It
rises through the adapted stack, peaks at the feature extraction layer, and declines slowly through the 16 frozen layers above it.}
\label{fig:seedgeom}
\end{figure}

\S\ref{sec:scaleout} reports that changing the seed gives almost no ensemble gain, whereas changing the readout or the backbone does. Using the 15 seeds of the main configuration, we show that adapters trained from different seeds compute nearly the same function, so their predictions have little independent content to average.

\paragraph{Weights.}
Updates from two seeds are nearly orthogonal in weight space: $\cos(\Delta W_i, \Delta W_j) = 0.026$, at a relative distance of $1.396 \approx \sqrt{2}$. This follows from the parameterization. $\Delta W = BA$ lies in the row space of $A$, which is drawn at random and changes little during training, so any two seeds are orthogonal regardless of what they compute. The update is not small: its largest singular value is $0.6$--$0.8$ of the frozen matrix's. Adapters should therefore be compared on activations
rather than weights.

\paragraph{Activations.}
Fig.~\ref{fig:seedgeom} compares each seed's change relative to the frozen backbone at increasing depth. On real inputs, a single module's output change agrees at 0.151. The composed feature change agrees at 0.355 after one layer, 0.595 after sixteen, and 0.860 at the feature extraction layer, where it is still $45\%$ of the feature norm. Agreement falls back to $0.74$--$0.82$ by the final layer (Fig.~\ref{fig:seedgeom}b). The rise is not explained by independent per-layer noise averaging out: a model of independent per-layer noise, fitted at layer 1, predicts $0.82$--$0.87$ by layer 8, against an observed $0.43$--$0.54$.

\paragraph{Readout and prediction.}
Pairing an adapter with another seed's jointly trained readout costs $0.0012$, and with another seed's re-fit readout $0.0024$, both inside the seed spread. Under one shared readout, the prediction change is $0.418$ of the prediction norm and $r=0.934$ correlated across seeds, with a shared fraction of $0.939$. Each seed's deviation from the consensus is therefore $0.418\sqrt{1-0.939} = 0.104$ of a prediction. The ensemble analysis measures this member-specific component independently, from the members' own predictions, and finds that it correlates with the fMRI target at $-0.003$.

The seed-specific part of a prediction is thus small and carries no signal, and it is what averaging over seeds removes. This is why seed-only pools saturate by ${\approx}12$ members while a member with a different readout or backbone still adds accuracy (App.~\ref{app:roster}).

\section{Additional null results}
\label{app:negatives}

We summarize the design choices we explored that did not improve the performance.

\paragraph{Gated-MLP readout.}
After the re-fit it scores $+0.0007$ over the linear readout, inside the 0.00077 seed spread.

\paragraph{Cross-attention over pooled features.}
Cross-attention over the same 894 frame-grid vectors that the linear readout uses gives $+0.00030$ [$+0.00017$, $+0.00045$], inside the seed spread. The gain of the cross-attention readout comes from the larger token set, not from attention itself.

\paragraph{Raw token sequence.}
Widening the token set from ${\sim}3.2$k to the full 19{,}099-token sequence, including every vision patch token, gives no gain ($+0.00000$, $-0.00075$ and $+0.00009$ over three seed-matched pairs).

\paragraph{Training-data subsets.}
Dropping whole stimuli so that two members share $44.1\%$ of their training stimuli instead of $99.3\%$ changes their residual correlation by only $+0.00006$. Training on random fractions of the data lowers accuracy monotonically (0.8/0.6/0.4 of the data $\to$ 0.3236/0.3161/0.3109).

\paragraph{Per-subject adapters.}
Restricting each adapter to one subject changes the score by $-0.0019$ [$-0.0022$, $-0.0017$] relative to the shared adapter, replicated at a second seed. A compute-matched control, with the same extra training but no subject restriction, loses by a similar amount, so the loss comes from continued training rather than from per-subject specialization. 

\section{Shift-target control}
\label{app:shift}

The frozen-backbone control of Tab.~\ref{tab:ablation} shows that the adapter's $+0.0356$ does not come from readout training. It does not show that the gain depends on the stimulus. An adapter could in principle profit from any fMRI-shaped supervision, for example the parcels' spatial covariance, their autocorrelation, or the per-run scale. We test this with a control experiment: we train the same configuration on targets whose stimulus--response pairing is broken, then re-fit the readout in closed form on correctly aligned training data and score it on the usual validation windows.

\paragraph{Setup.} For each subject, the training runs are concatenated in a seeded random order, rolled by a random offset (between 7.3 and 54.5 hours of scan time), and cut back into runs of their original lengths. Every target is still a real response of that subject. Window geometry, per-run z-scoring and within-run autocorrelation are preserved (mean lag-1 autocorrelation changes by less than 0.01), which makes this a stricter control than shuffling timepoints. Only the pairing with the stimulus is broken. Because each \emph{movie10} film is scanned twice, we check that no shifted target comes from a stimulus span that overlaps the window the model encodes. The re-fit and all evaluations use aligned data only.

The shifted adapter configuration also trains its per-modality projections on misaligned targets. Its matched control is therefore a second run at the same shift with the adapter's learning rate set to zero, so the two configurations differ only in the adapter.

\begin{table}[ht]
\caption{\textbf{Shifted-target negative control.} Friends s06 subject-mean after the closed-form re-fit on aligned training data, one run per configuration. Differences are paired per parcel with a functional-block bootstrap interval; the last column counts the parcels on which the first configuration scores higher.}
\label{tab:shift}
\centering
\footnotesize
\setlength{\tabcolsep}{5pt}
\begin{tabular}{llr}
\toprule
Configuration & Training targets & s06 \\
\midrule
LoRA\,+\,linear (ours) & aligned & 0.3263 \\
Frozen backbone & aligned & 0.2907 \\
Frozen backbone & shifted & 0.2617 \\
LoRA\,+\,linear & shifted & 0.1874 \\
\bottomrule
\end{tabular}

\vspace{6pt}
\begin{tabular}{lrr}
\toprule
Comparison & $\Delta$ [95\% CI] & parcels \\
\midrule
LoRA vs.\ frozen, both shifted & $-0.0733$ [$-0.0799$, $-0.0675$] & 4 / 1{,}000 \\
shifted LoRA vs.\ aligned frozen & $-0.1032$ [$-0.1111$, $-0.0959$] & 1 / 1{,}000 \\
shifted vs.\ aligned frozen & $-0.0299$ [$-0.0319$, $-0.0280$] & 0 / 1{,}000 \\
aligned vs.\ shifted LoRA & $+0.1389$ [$+0.1300$, $+0.1484$] & 1{,}000 / 1{,}000 \\
\bottomrule
\end{tabular}
\end{table}

\paragraph{Results.} Trained on shifted fMRI, the adapter degrades the features. It scores $0.0733$ below its matched control and loses on 996 of 1{,}000 parcels (Tab.~\ref{tab:shift}), an effect about 95 times the seed spread of the aligned configuration (0.00077). None of the $+0.0356$ survives once the stimulus--response pairing is removed, so the gain does not come from the marginal statistics of fMRI. The degradation is expected: an adapter that minimizes squared error against targets unrelated to its input moves the backbone's features toward the noise, and a re-fit of the readout alone cannot undo that. Misaligned training of the projections alone has the same effect at a smaller size ($-0.0299$, on every parcel).

\paragraph{Sanity check.} Before the re-fit, the shift-trained readouts score 0.0025 (adapter) and 0.029 (frozen) on aligned validation, down from 0.261 at step 0. Any stimulus-locked signal leaking into the shifted targets would have kept these scores above zero.

\paragraph{Scope.} Because misaligned training is harmful, the shifted adapter configuration does not estimate what the backbone is worth without a useful adapter; the aligned frozen row does. Each shifted configuration is a single run at a single shift, which already gives big enough effect sizes. 

\section{Does the adapter transfer to an unseen subject?}
\label{app:loso}

The shared adapter is trained on all four subjects, so the main results do not tell whether it learns about the stimulus or about these four people. We test this with leave-one-subject-out training. The main configuration is trained with the loss restricted to three subjects. The fourth subject's readout is then solved in closed form on the adapted features (the same re-fit as in \S\ref{sec:ablation}), and that subject is compared against itself under the frozen backbone and under adapters that did see it. There are four configurations, one per held-out subject, each with one seed and all other settings unchanged.

\begin{table}[htbp]
\caption{\textbf{Transfer to an unseen subject.}
Validation score per subject, with the readout solved in closed form, under:
the frozen backbone; an adapter trained on the \emph{other three} subjects
only (bold, the transfer); the mean of the three 3-subject adapters that
\emph{did} train on the subject (in-sample); and the shared 4-subject
adapter. The last three columns compare the held-out score with each of the
others.}
\label{tab:loso}
\centering
\footnotesize
\setlength{\tabcolsep}{4pt}
\begin{tabular}{lrrrrrrr}
\toprule
Held-out & frozen & \emph{3-subj., held out} & 3-subj., in-sample & 4-subject & vs.\ frozen & vs.\ in-sample & vs.\ 4-subj. \\
\midrule
sub-01 & 0.2875 & \textbf{0.3210} & 0.3215 & 0.3229 & +0.0335 & -0.0005 & -0.0019 \\
sub-02 & 0.2968 & \textbf{0.3286} & 0.3286 & 0.3301 & +0.0317 & -0.0000 & -0.0015 \\
sub-03 & 0.3130 & \textbf{0.3485} & 0.3490 & 0.3497 & +0.0355 & -0.0005 & -0.0012 \\
sub-05 & 0.2734 & \textbf{0.3034} & 0.3014 & 0.3025 & +0.0301 & +0.0021 & +0.0010 \\
\bottomrule
\end{tabular}
\end{table}

\paragraph{Results.} The adapter's gain transfers to an unseen subject (Tab.~\ref{tab:loso}). A held-out subject gains $+0.0301$ to $+0.0355$ over the frozen backbone, on 995 to 998 of 1{,}000 parcels. The remaining deficit against the four-subject adapter ($-0.0019$ to $+0.0010$) reflects the amount of supervision rather than the subject being unseen. Each three-subject adapter also scores its three in-sample subjects $0.0007$ to $0.0015$ below the four-subject adapter (paired block bootstrap, all intervals clear of zero). Within a configuration, the held-out subject scores the same as the in-sample ones ($-0.0005$ to $+0.0021$; sub-05 is better when held out).

Supervision from other people adapts the backbone as well as a subject's own data does. This is also consistent with the negative result for per-subject adapters (App.~\ref{app:negatives}): if three other subjects' data provide a subject's full gain, the gain does not depend on the subject’s own data. This test does not cover transfer across scanners, protocols, or datasets, which App.~\ref{app:nndb} addresses. 

\section{Does the adapter transfer to a new dataset?}
\label{app:nndb}

Leaving a subject out (App.~\ref{app:loso}) changes the person but keeps the site, scanner, preprocessing, repetition time and stimulus fixed. The Naturalistic Neuroimaging Database \citep[NNDb;][]{aliko2020nndb} changes all of them. Each participant watched one full-length feature film at a repetition time of 1\,s. The data are distributed in volumetric MNI space, so they can be parcellated into the same 1{,}000 Schaefer parcels our readout predicts. We use the 20 participants who watched the same film (5{,}470 volumes each) and score the frozen and LoRA models of Tab.~\ref{tab:main} without retraining the backbone or adapter.

\paragraph{Stimulus alignment.} NNDb does not distribute the films, and our copy differs from the scanned one in rate and offset. The rate difference is a PAL speedup (a 23.976\,fps master played at 25\,fps), so local timestamps map to scanner time by the ratio $960/1001$. We estimate the offset from NNDb's word-level annotations of the film. We verify the alignment in four ways. Re-fitting the clock on the retimed stimulus gives a rate of 1.00002 and a residual offset of $-0.075$\,s. The speech envelope of our media peaks in auditory cortex within 0.09\,s of the lag at which NNDb's word annotations peak. Frame-to-frame visual change peaks in V1 and MT. Finally, a lag scan of both models' predictions against the brain peaks at zero.

\paragraph{Readouts.} The \emph{transferred} readout applies the Algonauts-trained weights unchanged, averaging the four subject-specific rows since NNDb participants have none. This matches the validation setting but tests two things at once: whether the features transfer and whether a readout fitted to other people does. The \emph{re-fitted} readout separates the two. It keeps the backbone and projections from Algonauts and solves a fresh ridge
readout per NNDb participant, with five-fold cross-validation over contiguous blocks of film time and the penalty chosen on a holdout inside each training fold. All NNDb scores are computed on the model's 1.49\,s grid.

\paragraph{Noise ceiling.} Correlations measured on noisier data shrink for both models by the same factor. A raw gain in $r$ is therefore not comparable across datasets, and a relative gain can be inflated by noise. We report each model as a percentage of each dataset's own noise ceiling. The usual estimate, leave-one-out inter-subject correlation, correlates a subject with the mean of the others, and that mean is itself noisy. If a shared signal explains a fraction $f$ of the variance, it converges to $f/\sqrt{f + (1-f)/k}$ with $k$ other subjects, whereas an ideal stimulus model reaches $\sqrt{f}$. We invert the first expression for $f$ per parcel and report against $\sqrt{f}$. The correction matters when $k$ is small: against the raw estimate, the LoRA model would score 179\% of the ``ceiling'' on Algonauts, where $k=3$.

\begin{table}[H]
\caption{\textbf{Transfer to a second dataset.} The same LoRA and
frozen-backbone models on Algonauts validation and on NNDb (20 participants,
one film; different site, scanner, preprocessing and TR).
\emph{Transferred} applies the Algonauts readout unchanged; \emph{re-fitted}
solves a fresh readout per NNDb participant under cross-validation over
contiguous film time. $r$ columns are subject means over all 1{,}000 parcels.
Ceiling columns give $r$ as a percentage of each dataset's noise ceiling,
over the 922 (Algonauts) and 816 (NNDb) parcels with an estimable ceiling,
with a 95\% parcel-block bootstrap on the gain. Paired over the 800 parcels
both datasets admit, transferred minus Algonauts is $+0.55$ [$+0.01$,
$+1.06$] points.}
\label{tab:nndb}
\centering
\footnotesize
\setlength{\tabcolsep}{4pt}
\begin{tabular}{lrrrrrr}
\toprule
& \multicolumn{3}{c}{subject-mean $r$} & \multicolumn{3}{c}{\% of noise ceiling} \\
\cmidrule(lr){2-4}\cmidrule(lr){5-7}
& frozen & LoRA & gain & frozen & LoRA & gain, points [95\% CI] \\
\midrule
Algonauts, validation & 0.2907 & 0.3263 & +0.0356 & 81.3 & 92.0 & +10.7 [+10.0, +11.4] \\
NNDb, readout transferred & 0.0991 & 0.1223 & +0.0232 & 43.5 & 54.6 & \textbf{+11.1} [+10.5, +11.6] \\
NNDb, readout re-fitted & 0.0914 & 0.1252 & +0.0338 & 40.3 & 55.7 & +15.4 [+14.6, +16.2] \\
\bottomrule
\end{tabular}
\end{table}

\paragraph{Transferred readout.} The adapter's gain is $+11.1$ points of NNDb's ceiling, compared with $+10.7$ on Algonauts validation (Tab.~\ref{tab:nndb}). Paired over the parcels both datasets admit, the difference is $+0.55$ points [$+0.01$, $+1.06$]. The adapter improves 989 of 1{,}000 parcels and all 20 participants. In raw $r$ the gain looks smaller ($+0.0232$ against $+0.0356$) and in relative terms larger ($23\%$ against $12\%$). Both differences reflect where each dataset's frozen baseline sits relative to its ceiling, not the adapter.

\paragraph{Re-fitted readout.} The re-fitted readout shows a larger gain, $+15.4$ points of the ceiling and $+0.0338$ in $r$ ([$+0.0293$, $+0.0386$] by parcel-block bootstrap; 990 of 1{,}000 parcels, all 20 participants). The difference from the transferred readout lies almost entirely on the frozen side: re-fitting changes the LoRA model by $+0.0029$ and the frozen model by $-0.0077$. A readout solved on four-fifths of one film is worse than the Algonauts readout for the frozen features but not for the adapted ones. This is not only a matter of data quantity. On a learning curve over training windows, the LoRA features with half of the data (0.105) already exceed the frozen features with all of it (0.091), and the gap widens with more data. Choosing the penalty per parcel instead of globally lowers both configurations and leaves the gain at $+0.033$. We interpret the larger gain as the adapted features being easier to read out, not as a larger effect.

\begin{figure}[t]
\centering
\includegraphics[width=\linewidth]{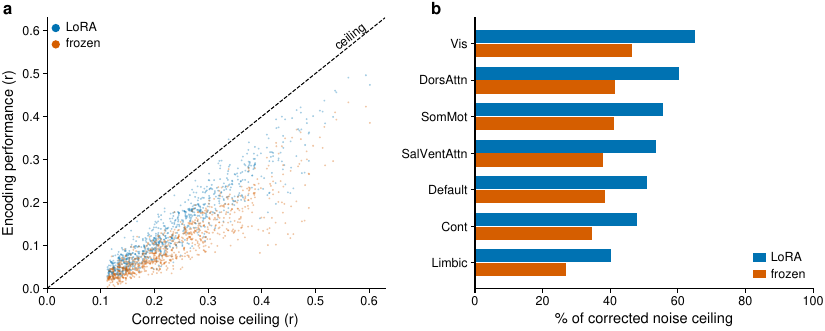}
\caption{\textbf{Performance against the noise ceiling on NNDb.} 20 participants, readout re-fitted per participant on the fixed backbone, over the 816 parcels with an estimable ceiling. \emph{(a)} Subject-mean $r$ per parcel against the corrected noise ceiling (dashed line). \emph{(b)} Mean percentage of the ceiling per Yeo network.}
\label{fig:nndb}
\end{figure}

\paragraph{Spatial pattern.} Across parcels, the gain map on NNDb correlates 0.620 with the gain map on Algonauts. Both maps are reliable (split-half reliability 0.891 on Algonauts and 0.940 on NNDb), giving a disattenuated correlation of 0.677. About $46\%$ of the spatial pattern is therefore shared, and the rest is dataset-specific structure rather than noise. At the network level (Fig.~\ref{fig:nndb}b), the adapter improves every network. The adapted model is closest to the ceiling in the visual and dorsal attention networks and furthest in the limbic network.

\section{Transfer to a new subject with little data}
\label{app:fewshot}

Apps.~\ref{app:loso} and~\ref{app:nndb} solve a new subject's readout on all of that subject's data, or apply the population readout zero-shot. In practice, a new participant may bring minutes to a few hours of recordings. We therefore ask how the readout should be fitted from little data, and whether the adapter's gain persists in that regime.

\paragraph{Setup.} We keep everything fixed except the new subject's readout row: the backbone, the adapter, the per-modality projections, and the training subjects' data are all kept constant. On Algonauts we use the four 3-subject adapters of App.~\ref{app:loso}, so the held-out subject's fMRI never reached the adapter. From that subject's Friends training windows we draw $K$ whole windows (five nested random draws per $K$, the same windows for every estimator), fit its readout row in closed form, and score it on season 6 as in Tab.~\ref{tab:loso}. We compare the following three ways of training the readout (Tab.~\ref{tab:fewshot}):
\begin{itemize}[noitemsep, topsep=0pt, partopsep=0pt, parsep=0pt, leftmargin=20pt]
    \item \textbf{Ridge from scratch}: the closed-form re-fit of \S\ref{sec:method}, on the $K$ windows alone.
    \item \textbf{Anchored ridge}: ridge on the residual $Y - X\bar W$, where $\bar W$ is the mean of the training subjects' rows, so the estimate shrinks toward the population readout instead of toward zero. With no data, it is equivalent to the population readout.
    \item \textbf{Low-rank deviation}: $\bar W + AB^\top$, with $B$ fixed to the top 16 principal components of the training subjects' residuals under $\bar W$, and only $A$ fitted. This is a closed-form analogue of the shared and subject-specific decomposition of RABBiT \citep{moussa2026rabbit}.
\end{itemize}
Each estimator uses one penalty for all parcels, chosen by cross-validation over whole calibration windows and never on season 6. For the last two methods, the penalty tuning grid includes $\alpha = \infty$, so they can fall back to the population readout. The frozen-backbone comparison repeats every fit on the 3-subject frozen-backbone models of Tab.~\ref{tab:loso}.

\begin{table}[ht]
\caption{\textbf{Calibrating a new subject with little data.} Validation $r$ of
a subject whose fMRI was never used for training the adapter, with its readout fitted on $K$ windows (149\,s each) of its own data, shown in minutes. Bold marks the best estimator at each budget. The last row is the gain of the adapted model over the frozen backbone under anchored ridge, which beats it for every subject and draw (20/20) and every NNDb participant (20/20) at every budget. 
\emph{Top:} the four 3-subject adapters of Tab.~\ref{tab:loso}, mean over the four held-out subjects and five random draws of the calibration windows; \emph{all} is the held-out column of Tab.~\ref{tab:loso}, solved on all of that subject's training data.
\emph{Bottom:} the LoRA\,+\,linear model of  Tab.~\ref{tab:main} on NNDb, 20 participants, a fixed 7-window test segment and a contiguous calibration run, averaged over placements at the start and end of the film. Its 0\,min scores therefore differ from Tab.~\ref{tab:nndb}, which scores the whole film.
}
\label{tab:fewshot}
\centering
\footnotesize
\setlength{\tabcolsep}{4pt}
\begin{tabular}{lrrrrrrrr}
\toprule
& \multicolumn{8}{c}{calibration data from the new subject (min)} \\
\cmidrule(lr){2-9}
Algonauts, held-out subject & 0 & 5 & 10 & 20 & 40 & 79 & 159 & all \\
\midrule
Ridge from scratch & -- & 0.1495 & 0.1967 & 0.2273 & 0.2493 & 0.2687 & 0.2869 & 0.3254 \\
Low-rank deviation ($r{=}16$) & 0.2574 & 0.2572 & 0.2578 & 0.2591 & 0.2603 & 0.2621 & 0.2642 & -- \\
Anchored ridge & 0.2574 & \textbf{0.2597} & \textbf{0.2641} & \textbf{0.2718} & \textbf{0.2789} & \textbf{0.2889} & \textbf{0.2997} & -- \\
\midrule
LoRA $-$ frozen (anchored) & +0.0236 & +0.0254 & +0.0260 & +0.0272 & +0.0273 & +0.0298 & +0.0326 & +0.0326 \\
\bottomrule
\end{tabular}

\vspace{6pt}
\begin{tabular}{lrrrrrr}
\toprule
& \multicolumn{6}{c}{calibration data from the new participant (min)} \\
\cmidrule(lr){2-7}
NNDb, new dataset & 0 & 5 & 10 & 20 & 40 & 70 \\
\midrule
Ridge from scratch & -- & 0.0511 & 0.0725 & 0.0902 & 0.1075 & 0.1180 \\
Anchored ridge & 0.1176 & \textbf{0.1170} & \textbf{0.1220} & \textbf{0.1245} & \textbf{0.1290} & \textbf{0.1326} \\
\midrule
LoRA $-$ frozen (anchored) & +0.0215 & +0.0212 & +0.0237 & +0.0257 & +0.0288 & +0.0308 \\
\bottomrule
\end{tabular}
\end{table}

\paragraph{Anchoring is powerful in the low-data regime.} Ridge from scratch needs about 80\,min of the new subject's data merely to match the population readout ($0.2687$ against $0.2574$). Anchored ridge beats it for all 20 subject--draw pairs at every budget. It improves on the population readout from the first minutes ($+0.0022$ at 5\,min, 17/20; $+0.0067$ at 10\,min, 19/20) and for all 20 pairs from 20\,min on: $+0.0143$ at 20\,min and $+0.0423$ at 159\,min. These close $21\%$ and $62\%$ of the gap to the subject's own full-data readout.

\paragraph{The low-rank deviation adds little.} Restricting the subject's correction to a basis from the other subjects gains only $+0.0029$ at 40\,min and $+0.0067$ at 159\,min over the population readout, and anchored ridge beats it for all 20 pairs from 20\,min on. Here, the gain from calibration comes from anchoring to the shared readout, not from the low rank. This test differs from RABBiT itself: RABBiT learns its decomposition end to end and per region, from more subjects than our three, whereas we fix the basis in closed form. It therefore tests the low-rank restriction at our subject count, not the RABBiT method.

\paragraph{The adapter's gain persists at every budget.} Under anchored ridge, the adapted model beats the frozen backbone for every subject and draw at every budget. The gain is $+0.0236$ with no data from the new subject, and it rises to $+0.0326$ at 159\,min, the same as with all of the subject's data. The adapted features also gain more from the same calibration data ($+0.0423$ against $+0.0333$ at 159\,min).

\paragraph{On a new dataset NNDb.} On NNDb we apply the same estimators to the LoRA\,+\,linear model of Tab.~\ref{tab:main} and its frozen-backbone counterpart, both as trained on Algonauts, with the mean of the four Algonauts rows as the population readout. Each participant's film is split into a fixed 7-window test segment and a contiguous calibration run, separated by one window. We average the two placements of this split, at the start and at the end of the film. Ridge from scratch needs the full 70\,min to match the transferred readout ($0.1180$ against $0.1176$). Anchored ridge improves on it by $+0.0113$ at 40\,min (all 20 participants) and $+0.0150$ at 70\,min (19 of 20). The adapter's gain over the frozen backbone widens from $+0.0215$ to $+0.0308$, for all 20 participants at every budget. The low-rank deviation is not run on NNDb.

\section{Where in the brain the adapter helps}
\label{app:anatomy}

\begin{table}[H]
\caption{\textbf{Gain from the adapter by region.} Validation $r$ with
and without LoRA in the same configuration (the frozen arm trains the readout
with the same rates and steps), over the $n$ parcels of each region: the
seven Yeo networks, then ten ROIs defined from parcel centroids. Arms are
paired per parcel; gain intervals are functional-block bootstraps over
parcels, and the last column is the gain relative to the frozen score. One
run per arm; averaging the LoRA map over 15 seeds leaves both orderings
unchanged.}
\label{tab:anatomy}
\centering
\footnotesize
\setlength{\tabcolsep}{5pt}
\begin{tabular}{lrrrrr}
\toprule
region & $n$ & frozen & LoRA & gain [95\% CI] & rel.\ \% \\
\midrule
\multicolumn{6}{l}{\emph{Yeo 7 networks}} \\
DorsAttn & 122 & 0.330 & 0.380 & +0.0502 [+0.0468, +0.0533] & 15 \\
Vis & 162 & 0.398 & 0.438 & +0.0401 [+0.0364, +0.0440] & 10 \\
Cont & 129 & 0.279 & 0.317 & +0.0379 [+0.0347, +0.0412] & 14 \\
SalVentAttn & 121 & 0.241 & 0.277 & +0.0352 [+0.0319, +0.0397] & 15 \\
Default & 212 & 0.318 & 0.351 & +0.0330 [+0.0306, +0.0355] & 10 \\
SomMot & 194 & 0.234 & 0.265 & +0.0307 [+0.0288, +0.0336] & 13 \\
Limbic & 60 & 0.131 & 0.146 & +0.0147 [+0.0121, +0.0198] & 11 \\
\midrule
\multicolumn{6}{l}{\emph{coordinate-defined ROIs}} \\
MT+ / lat. occipitotemporal & 10 & 0.490 & 0.545 & +0.0547 [+0.0468, +0.0598] & 11 \\
IPS / superior parietal & 66 & 0.294 & 0.348 & +0.0543 [+0.0499, +0.0587] & 18 \\
TPJ / supramarginal & 46 & 0.321 & 0.370 & +0.0494 [+0.0445, +0.0538] & 15 \\
early visual (V1/V2) & 54 & 0.412 & 0.455 & +0.0433 [+0.0401, +0.0493] & 10 \\
precuneus / PCC (DMN) & 34 & 0.307 & 0.347 & +0.0399 [+0.0340, +0.0436] & 13 \\
IFG / frontal operculum & 59 & 0.260 & 0.295 & +0.0350 [+0.0327, +0.0378] & 13 \\
somatomotor strip & 61 & 0.171 & 0.202 & +0.0316 [+0.0285, +0.0366] & 19 \\
early auditory (HG/PT/STG) & 35 & 0.403 & 0.434 & +0.0307 [+0.0275, +0.0322] & 8 \\
STS / lateral temporal & 42 & 0.408 & 0.438 & +0.0299 [+0.0278, +0.0326] & 7 \\
medial PFC (DMN) & 39 & 0.194 & 0.218 & +0.0246 [+0.0168, +0.0318] & 13 \\
\bottomrule
\end{tabular}

\end{table}

\begin{figure}[t]
\centering
\includegraphics[width=\linewidth]{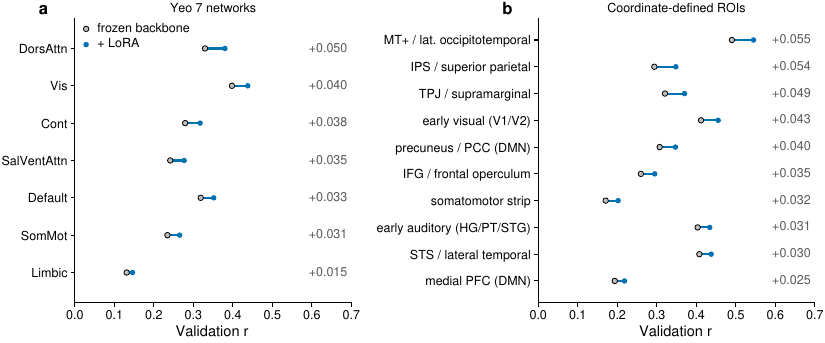}
\vspace{-20pt}
\caption{\textbf{Validation $r$ by region, with and without the adapter.}
Frozen backbone and the same configuration with LoRA, over the $n$ parcels of each region (Tab.~\ref{tab:anatomy}); bars show the gain, printed at the right. \emph{(a)} The seven Yeo networks. \emph{(b)} Ten coordinate-defined ROIs. One run per configuration, paired per parcel; every gain's 95\%
functional-block bootstrap interval excludes zero.}
\vspace{-18pt}
\label{fig:anatomy}
\end{figure}

The first two configurations of Tab.~\ref{tab:ablation} share a configuration and differ only in LoRA, so their per-parcel scores are paired and their difference isolates the adapter. The adapter adds $+0.0356$ on average and improves every one of the 1{,}000 parcels. The four subjects' gain maps agree at a mean pairwise $r$ of $0.673$, and the orderings below are unchanged when the LoRA map is averaged over 15 seeds. Tab.~\ref{tab:anatomy} and Fig.~\ref{fig:anatomy} break the scores down by the seven Yeo networks and by ten ROIs defined from parcel centroids.

\paragraph{Regional gains.} Every region improves, with the largest gains in parietal and visual cortex. Among the networks, dorsal attention gains most ($+0.050$ on a frozen $0.330$) and limbic least ($+0.015$ on $0.131$). Among the ROIs, MT+ ($+0.055$), the intraparietal sulcus ($+0.054$) and the temporoparietal junction ($+0.049$) lead, and medial prefrontal cortex trails ($+0.025$). Relative to the frozen score, which varies threefold across regions, the ordering changes: the somatomotor strip and the intraparietal sulcus gain $19\%$ and $18\%$, while MT+ and early visual cortex gain $11\%$ and $10\%$.

\paragraph{Language regions.} Early auditory cortex and the STS/lateral temporal cortex, where an adapted language model might be expected to help most, have the smallest relative gains: $+0.031$ ($8\%$ of a frozen $0.403$) and $+0.030$ ($7\%$ of $0.408$). Both already have among the highest frozen scores. There is also no hemispheric asymmetry (LH $-$ RH $-0.0010$ [$-0.0030$, $+0.0008$]).

\paragraph{Reliability.} Across all 1{,}000 parcels the gain correlates $+0.379$ with the frozen score, but only $-0.016$ across the $757$ parcels whose \emph{movie10} test--retest reliability exceeds $0.1$. Poorly measured parcels have both low frozen scores and small gains, because measurement noise attenuates both. The gain is therefore not simply larger where the model was already good.

\section{Input-level modality attribution}
\label{app:modality}

The attribution of \S\ref{sec:interpret} is computed on the LoRA\,+\,linear model of Tab.~\ref{tab:main} and on its frozen-backbone counterpart, over the same 250 validation windows. A modality is removed by replacing its input embeddings to the language model with their mean over training windows. For video, this covers the visual embeddings and their multi-scale copies (which would otherwise reintroduce video content in the first layers); for audio, the audio tower's output; for text, the transcript tokens, leaving template tokens untouched. Token counts, positions and timing are unchanged, so only stimulus content is removed. We evaluate all $2^3$ subsets per window, which makes the three Shapley values exact. The full-subset predictions match the
run's evaluation output exactly, and the Shapley values sum to the prediction
(residual $<10^{-12}$).

\begin{figure}[t]
\centering
\includegraphics[width=\linewidth]{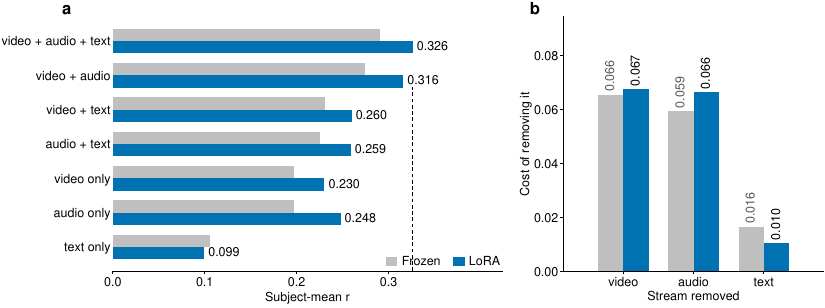}
\caption{\textbf{All subsets of the three input streams.} Subject-mean Pearson $r$ on the 250 validation windows used for attribution, with absent streams replaced by their training-mean embeddings. The top bar of \emph{(a)} is the full model (dashed line). \emph{(a)} Any single stream recovers most of the full score, and removing one costs far less than that stream achieves alone: the streams are largely redundant. \emph{(b)} Cost of removing each stream, before and after adaptation. The video cost barely changes, the audio cost grows, and the transcript cost falls by more than a third. This removes a stream from a \emph{trained} model and differs from the no-audio configuration of Tab.~\ref{tab:ablation}, which is trained without audio throughout.}
\label{fig:coalitions}
\end{figure}

\paragraph{Modality subsets.} Fig.~\ref{fig:coalitions} shows the subsets themselves in addition to the decomposition. The streams are largely redundant: video alone recovers $71\%$ of the full score and audio alone $76\%$. Removing video or audio costs about the same ($0.0675$ and $0.0665$), but after adaptation audio alone outscores video alone ($0.2478$ against $0.2303$), whereas the frozen backbone ties them ($0.1974$ and $0.1975$). Removing the transcript costs little ($0.0103$), although it reaches $0.0994$ on its own, because the audio carries the same information.

Adaptation sharpens this. Removing the transcript costs $0.0164$ with the frozen backbone and $0.0103$ after LoRA. As single streams, audio gains $+0.0503$ from adaptation and video $+0.0329$, while the transcript loses $0.0065$. The adapted model thus draws from the speech audio what the frozen model obtained from the transcript, matching the Shapley shares of \S\ref{sec:interpret}.

These removal costs are not comparable to the \emph{No audio} row of Tab.~\ref{tab:ablation} ($0.3026$). That configuration is trained without audio from the start, so its adapter and readout compensate for the missing stream, and it loses $0.0232$ against its matched control. Removing audio from a model trained with it costs $0.0665$.

\paragraph{Cross-stream contribution.} The readout takes one hidden state per stream every $0.5$\,s, at a video, an audio and a transcript token position. By feature extraction layer 32 the decoder has mixed the sequence, so the content at a position need not be the content that entered there. Running the Shapley decomposition within each position group gives the $3\times3$ flow of Fig.~\ref{fig:modality}d, from stimulus modality (left) to the token type whose hidden state carries it (right). \emph{Layout} is the contribution of the token arrangement with all content removed.

\begin{figure}[t]
\centering
\includegraphics[width=\linewidth]{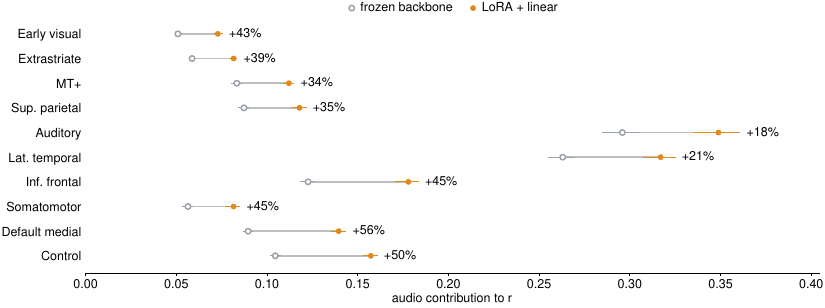}
\caption{\textbf{Audio contribution by region, before and after adaptation.}
Subject-mean contribution of the audio stream to encoding accuracy for the
frozen and LoRA\,+\,linear models, with episode block-bootstrap intervals;
labels give the relative change. Auditory cortex gains least; the gain is
largest in regions that relied least on audio before adaptation.}
\label{fig:audio-region}
\end{figure}

Adaptation nearly doubles cross-stream contribution, from $8.9\%$ [$8.6$, $9.2$] of the prediction with the frozen backbone to $15.3\%$ [$14.9$, $15.6$] after LoRA (paired difference $+6.4$ points [$+6.1$, $+6.6$]). The flow is asymmetric: video information read at audio positions accounts for $7.5\%$, against $3.9\%$ in the other direction (frozen: $5.0\%$ and $1.1\%$). Audio positions carry $35.1\%$ of the frozen prediction and $45.8\%$ of the adapted one, while transcript positions fall from $22.5\%$ to $14.2\%$. This is the same shift toward audio seen in cortex (Fig.~\ref{fig:modality}c, Fig.~\ref{fig:audio-region}), here at the token level. Text input information read at audio output positions rises from $0.5\%$ to $2.0\%$, and audio read at text positions falls from $1.8\%$ to $1.5\%$. Negative flows are clipped in the diagram; they amount to $0.2\%$ of the frozen prediction and none of the
adapted one. The percentages in the text are exact signed means.

\paragraph{Statistics.} We report two statistics per parcel. The \emph{share} $\mathrm{cov}(\phi_m, \hat y)/\mathrm{var}(\hat y)$ describes what the prediction is made of and sums to one. The \emph{contribution to accuracy} $\mathrm{cov}(\phi_m, y)/(\sigma_{\hat y}\sigma_y)$ describes which part of it tracks the response and sums to $r$. The text quotes shares; maps meant to localize use the contribution, since a share is defined even where the model predicts poorly. Intervals are block bootstraps over validation episodes.

\paragraph{Robustness.} As an alternative baseline, we replace an absent modality's embeddings with those of a window from a different episode, which is in distribution per token but wrong in content. On the LoRA\,+\,linear model the two baselines give share maps correlated at $0.994$ and agree on the dominant modality for $98.2\%$ of parcels.

The frozen-to-adapted comparison of \S\ref{sec:interpret} is between two models, so we compare it with the agreement between two adapted seeds. Maps are compared over the 971 parcels that both models predict at $r \ge 0.1$. Two adapted seeds agree at $0.996$--$1.000$ on the contribution maps and on the dominant modality for $97.2\%$ of parcels. The frozen and adapted models agree at only $0.963$--$0.990$ and on $74.4\%$ of parcels, outside the seed range with bootstrap intervals clear of it. With the LoRA\,+\,cross-attention model, the dominant modality changes on fewer parcels ($8.0\%$ against $25.6\%$) but in the same direction, and cross-stream contribution rises from $11.9\%$ [$11.5$, $12.2$] to $15.8\%$ [$15.4$, $16.3$]. The reallocation is therefore a property of the adaptation, not of the readout.

\section{What the pool still misses}
\label{app:partition}

Validation correlation does not separate model error from missing signal and measurement noise. A residual mixes model error, missing signal and measurement noise, and the validation season has no test-retest to separate them. Two properties of our setup allow the separation. First, the member-specific part of each member's prediction is noise (it correlates with the target at $-0.003$, App.~\ref{app:seedgeom}), so the \emph{uniform} mean of the 261 members is the pool's shared function with that noise averaged out. Second, all four subjects watched the same 250 validation windows. Scanner and idiosyncratic noise cannot correlate across people, so a residual that is correlated \emph{across subjects} is stimulus-locked signal that the pool does not produce.

For subject $s$ and parcel $p$, let $y$ be the z-scored fMRI, $\hat y$ the pool mean rescaled by its least-squares gain, $e = y - \hat y$, and $t$ the mean of the other three subjects, over the 25{,}000 validation TRs. Then
\[
\underbrace{\mathrm{cov}(y_s, y_t)}_{\text{ISC}}
 = \underbrace{\mathrm{cov}(\hat y_s, \hat y_t)}_{\text{explained}}
 + \underbrace{\mathrm{cov}(e_s, e_t)}_{\text{missed}}
 + \underbrace{\mathrm{cov}(\hat y_s, e_t) + \mathrm{cov}(e_s, \hat y_t)}_{\text{cross}} .
\]
The left side is the inter-subject covariance, the stimulus-locked variance available to any stimulus-computable model. The ``missed'' term is the part of it the pool does not explain. The cross term is $1.3\%$ of ISC.

\begin{table}[H]
\caption{\textbf{What the pool still misses.} The cross-subject partition
of the stimulus-locked (inter-subject) covariance on the validation episodes,
as a percentage of it: the part shared by the predictions (explained), the
part shared by the residuals (missed) and the cross terms, for three pools;
the interval is a functional-block bootstrap over parcels, and the last column
is the missed covariance in units of a unit-variance parcel.}
\label{tab:partition}
\centering
\footnotesize
\setlength{\tabcolsep}{5pt}
\begin{tabular}{lrrrrr}
\toprule
pool & val.\ $r$ & explained \% & \textbf{missed \%} & cross \% & missed \\
\midrule
main single model (LoRA + linear) & 0.3263 & 60.5 & 36.2 & 3.3 & 0.0480 \\
3-member mean & 0.3262 & 60.8 & 36.5 & 2.6 & 0.0485 \\
\textbf{261-member uniform mean} & 0.3401 & 66.2 & 32.5 [30.0, 35.3] & 1.3 & 0.0431 \\
\bottomrule
\end{tabular}

\end{table}

\paragraph{Results.}
The pool leaves $32.5\%$ [30.0, 35.3] of the cross-subject stimulus-locked variance unexplained (Tab.~\ref{tab:partition}). A single model leaves $36.2\%$, so averaging 261 models recovers 3.7 points. The four subjects agree to within 1.4 points. In variance units the pool misses 0.043 of a unit-variance parcel, against an ISC of 0.133.

This is a lower bound on what is missed relative to all stimulus-driven variance, because ISC does not count stimulus-driven signal that is subject-specific. Within a subject the pool explains variance equal to $94\%$ of ISC. Only 66\% points of ISC across subjects' predictions; the other $28$ are subject-specific stimulus-driven signal. Measured against captured plus missed variance, the missed share is $26\%$.

The same pool reaches $97\%$ of the ISC ceiling in correlation ($\sqrt{\mathrm{ISC}/\mathrm{var}}$), which corresponds to $94\%$ in variance. Distance to the ceiling is therefore better stated in variance.

\paragraph{Characterizing the residual.}
Nine low-level stimulus statistics (luminance, contrast, motion energy and cuts from the frames; RMS, envelope modulation and spectral centroid from the audio; words per TR and speech on/off from the transcript), lagged 0--9\,s, explain the shared residual at only $r = 0.027$ under episode-level cross-validation. The same features explain the pool's prediction at 0.48, so the pool already carries this information. Lagged copies of the pool's own prediction ($-4$ to $+4$ TRs) explain $0.048$, so the residual is not a latency error. It is also not low-dimensional: the top ten components hold $36\%$ of its variance, against $91\%$ for the prediction. Its spectrum
follows that of the ISC ($66\%$ of the missed covariance lies below 0.05\,Hz, against $70\%$ of ISC). The residual is thus slow, distributed, high-dimensional and stimulus-locked, which fits narrative or event-level structure better than a missing low-level feature. This agrees with \S\ref{sec:scaleout}: even the changes that most improved the pool left about a third of the shared variance unexplained.

\section{Reproducibility}
\label{app:config}

\paragraph{Hardware and precision.}
One NVIDIA A800-80GB per run, for training and extraction. The backbone runs in bfloat16 with FlashAttention-2 and gradient checkpointing. The vision patch embedding's 3-D convolution runs in float32, because cuDNN~9 executes it in bfloat16 roughly four orders of magnitude slower.

\paragraph{Optimisation.}
AdamW with default $\beta$, weight decay 0.01, gradient clipping at global norm 1.0. Batch size 1 window, no gradient accumulation; one window is 100 TRs for all four subjects, so a step carries 400 target rows. 1{,}429 steps (one epoch over the 1{,}429 training windows), 30 warm-up steps, cosine decay to zero. Learning rates: LoRA $6{\times}10^{-4}$, per-modality projection $6{\times}10^{-4}$, readout $3{\times}10^{-4}$. No run is resumed from a checkpoint.

\paragraph{LoRA.}
Rank 16, $\alpha = 32$, dropout 0, on \texttt{q\_proj}, \texttt{k\_proj}, \texttt{v\_proj} and \texttt{o\_proj} of every decoder layer (\texttt{model.layers.\textbackslash d+.self\_attn.(q|k|v|o)\_proj}). PEFT default initialization ($A$ Kaiming-uniform, $B$ zero), so training starts from the frozen model. 8.91M parameters. The decoder is truncated above the feature extraction layer, so layers past 32 are neither run nor adapted.

\paragraph{Full finetuning.}
The full-finetuning configurations of \S\ref{sec:main} train all $20.25$B parameters below the feature extraction layer with the settings above, except the backbone learning rate. We swept $3{\times}10^{-6}$ to $1{\times}10^{-4}$ in half-decades (constant schedule, 200 steps, 42-window selection subset). $1{\times}10^{-5}$ was best ($0.2753$, against $0.2684$ for the LoRA reference); $3{\times}10^{-6}$ was still improving and the two higher rates stalled or degraded. A step takes 19.8\,s against 12.0\,s for LoRA. Both configurations are single-seed, and the cross-attention model reuses the linear model's rate.

\paragraph{Clock and lag blocks.}
$\mathrm{TR} = 1.49$\,s and the feature grid is 2\,Hz, so 100 TRs is 149\,s and 298 frames. Hemodynamic alignment uses MIRAGE's fixed shift of 4.47\,s, i.e.\ an offset of $\mathrm{round}(4.47 \times 2) = 9$ frames: TR $j$'s block center is at frame $c_j = 9 + 2.98j$. Each of the 8 blocks is a (delay, width) pair $(\delta_b, w_b)$ in seconds, and block $b$ of TR $j$ averages frames $[\lceil c_j - (\delta_b + w_b)\,f \rceil, \lceil c_j - \delta_b f \rceil)$, clipped to the window, with a zero row where the block lies entirely before the window. The operator is one dense matrix shared by the training path and the closed-form solve.

Tab.~\ref{tab:lagblocks} lists the eight pairs. They tile $[0.5, 192.5)$\,s contiguously with widths $2, 2, 4, 8, 16, 32, 64, 64$\,s. Rebuilding the operator from these pairs reproduces the $800 \times 298$ matrix in the released checkpoint exactly. Because the deepest block reaches 192.5\,s back, beyond the 149\,s window, $158$ of the $800$ design rows ($19.8\%$) are zero and block 8 never reaches full width. Every non-zero row sums to one.

\begin{table}[t]
\caption{\textbf{Lag blocks.} Each block is a (delay, width) pair in seconds. Block $b$ of TR $j$ uniformly averages the feature frames in $[c_j - (\delta_b + w_b)f, c_j - \delta_b f)$, where $c_j$ is the TR's block center and $f = 2$\,Hz. A block that reaches before the window start is clipped, or is a zero row if it lies entirely before it. The last three columns count, over the 100 TRs of a window, the rows at full width, clipped, and zero (158 of 800 rows are zero).}
\label{tab:lagblocks}
\centering
\footnotesize
\begin{tabular}{rrrlrrr}
\toprule
Block & $\delta_b$ (s) & $w_b$ (s) & covers (s) & full & clipped & zero \\
\midrule
1 & 0.5 & 2.0 & $[0.5, 2.5)$ & 98 & 1 & 1 \\
2 & 2.5 & 2.0 & $[2.5, 4.5)$ & 99 & 1 & 0 \\
3 & 4.5 & 4.0 & $[4.5, 8.5)$ & 97 & 2 & 1 \\
4 & 8.5 & 8.0 & $[8.5, 16.5)$ & 92 & 5 & 3 \\
5 & 16.5 & 16.0 & $[16.5, 32.5)$ & 81 & 10 & 9 \\
6 & 32.5 & 32.0 & $[32.5, 64.5)$ & 60 & 21 & 19 \\
7 & 64.5 & 64.0 & $[64.5, 128.5)$ & 17 & 42 & 41 \\
8 & 128.5 & 64.0 & $[128.5, 192.5)$ & 0 & 16 & 84 \\
\bottomrule
\end{tabular}
\end{table}

\paragraph{Readout.}
Per-frame linear projection $2048 \to 512$ per modality (shared across frames), then the lag average, giving $8 \times 3 \times 512 = 12{,}288$ features per TR, then a per-subject linear map to 1000 parcels. Projecting before lag-averaging keeps empty blocks exactly zero. 52.30M parameters. Registered buffers (lag operator and centering statistics, 0.26M) are fixed and not counted.

\paragraph{Cross-attention readout.}
Replaces the lag operator with per-TR cross-attention over the language-model token sequence: 8 learned queries per modality, key/query width 128, a causal cutoff at each TR's block center, and a per-block additive bias (floored at $-12$) initialized to reproduce the lag operator's uniform weighting. It reads the \texttt{stream} token set, about 3.2k tokens per span: audio at its native 13\,Hz, all text tokens, and one mean-pooled vector per second of video. It adds 0.2M parameters and uses the same schedule. A stream with no token before a TR's cutoff is masked explicitly.

\paragraph{Ridge penalties.}
$\{10^3, 10^4, 10^5, 10^6, 10^7\}$, chosen per parcel and subject on the training-split holdout of App.~\ref{app:protocol}. For the main model 3{,}829 of 4{,}000 parcel-subject pairs select $10^4$, 110 select $10^5$, 60 select $10^3$ and one selects $10^6$.

\paragraph{Ensemble combiner.}
Per-parcel minimum-variance weights \citep{perrone1993disagree} on the members' residual covariance $S$: $w = S^{-1}\mathbf{1} / (\mathbf{1}^\top S^{-1}\mathbf{1})$, the minimizer of $w^\top S w$ subject to $\mathbf{1}^\top w = 1$. The constraint is free because per-parcel Pearson is invariant to positive rescaling. Weights may be negative. $S$ is shrunk toward its diagonal with $\lambda \in \{0.01, 0.03, 0.1, 0.3, 0.6, 0.9\}$, chosen on an inner episode split, never on the rows the weights are scored on. The outer split is in App.~\ref{app:protocol}.

\paragraph{Backbone revisions.}
Tab.~\ref{tab:revisions} gives the Hugging Face repository and commit for every backbone in Tab.~\ref{tab:backbones}. For \textsc{Phi-4-MM} and \textsc{LCO-Embedding-Omni}, upstream \texttt{main} has since moved past the commit we used.

\begin{table}[t]
\caption{\textbf{Backbone revisions.} Hugging Face repository and commit for every backbone in Tab.~\ref{tab:backbones}, as loaded by our runs.}
\label{tab:revisions}
\centering
\scriptsize
\setlength{\tabcolsep}{3pt}
\begin{tabular}{lll}
\toprule
Backbone & Hugging Face repository & Revision \\
\midrule
Phi-4-MM & \texttt{microsoft/Phi-4-multimodal-instruct} & \texttt{467a5d77dcfa} \\
Qwen2.5-Omni 3B & \texttt{Qwen/Qwen2.5-Omni-3B} & \texttt{f75b40e3da20} \\
Gemma 3n E4B & \texttt{unsloth/gemma-3n-E4B-it} & \texttt{45e9fb1dd0e3} \\
LCO-Omni 7B & \texttt{LCO-Embedding/LCO-Embedding-Omni-7B} & \texttt{3d38f58aae12} \\
MiniCPM-o & \texttt{openbmb/MiniCPM-o-4\_5} & \texttt{503e754207c9} \\
Qwen2.5-Omni 7B & \texttt{Qwen/Qwen2.5-Omni-7B} & \texttt{ae9e1690543f} \\
Qwen3-Omni 30B & \texttt{Qwen/Qwen3-Omni-30B-A3B-Thinking} & \texttt{2f443cfc4c54} \\
Nemotron Omni & \texttt{nvidia/Nemotron-3-Nano-Omni-30B-A3B-Reasoning-BF16} & \texttt{e5e9932441de} \\
Ming-flash 104B & \texttt{inclusionAI/Ming-flash-omni-2.0} & \texttt{6a2e1dec0706} \\
\bottomrule
\end{tabular}
\end{table}

\paragraph{What a rerun needs.}
The stimulus media cache at $336{\times}224$, the Algonauts 2025 fMRI, the backbone weights at the revisions in Tab.~\ref{tab:revisions}, and a readout initialization: a ridge solution on frozen features at the same context lead, with the lag basis of Tab.~\ref{tab:lagblocks}. Stage 1 is the 1{,}429-step run above; stage 2 is one forward-only pass that re-solves the readout in closed form.

\end{document}